\documentclass[11pt]{article}

\usepackage[preprint]{acl}

\usepackage{times}
\usepackage{latexsym}

\usepackage[T1]{fontenc}

\usepackage[utf8]{inputenc}

\usepackage{microtype}

\IfFileExists{inconsolata.sty}{\usepackage{inconsolata}}{}

\usepackage{graphicx}
\usepackage{changepage}
\usepackage{xspace}

\newcommand{\method}{\textsc{EvoRS}\xspace}
\newcommand{\rewarddag}{\textit{Reward-DAG}\xspace}

\usepackage{colortbl}
\usepackage{booktabs}
\usepackage{enumitem}
\usepackage{dblfloatfix}
\usepackage{placeins}
\usepackage[most]{tcolorbox}

\definecolor{mygray}{RGB}{226, 226, 226}
\definecolor{myred}{RGB}{252, 142, 142}
\definecolor{mygreen}{RGB}{147, 255, 143}
\definecolor{myblue}{RGB}{144, 155, 255}
\definecolor{myyellow}{RGB}{253, 253, 143}
\definecolor{mypurple}{RGB}{255, 142, 250}

\makeatletter
\ifacl@anonymize\else
\fi
\makeatother

\title{EvoRS: On-Policy Self-Evolution of Reward Systems for Open-Ended Reinforcement Learning}

\author{
  \textbf{Weiyuan Li\textsuperscript{1,2$\heartsuit$}},
  \textbf{Aili Chen\textsuperscript{2,3}},
  \textbf{Xintao Wang\textsuperscript{2,3}},
  \textbf{Yikai Zhang\textsuperscript{2,3}},
  \\
  \textbf{Qingqing Dong\textsuperscript{4}},
  \textbf{Jinghan Xu\textsuperscript{1,2}},
  \textbf{Hongru Hou\textsuperscript{1,2}},
  \textbf{Wenxuan Zhao\textsuperscript{5}},
  \textbf{Chengkun Lang\textsuperscript{5}},
  \\
  \textbf{Jun Gao\textsuperscript{5}},
  \textbf{Yuanli Guo\textsuperscript{2,6}},
  \textbf{Hongcheng Guo\textsuperscript{2,3}},
  \textbf{Yanghua Xiao\textsuperscript{2,3}},
  \textbf{Deqing Yang\textsuperscript{1,2$\spadesuit$*}},
  \\
  \textsuperscript{1}School of Data Science, Fudan University,
  \textsuperscript{2}Shanghai Key Laboratory of Data Science\\
  \textsuperscript{3}College of Computer Science and Artificial Intelligence, Fudan University\\
  \textsuperscript{4}College of Cryptology and Cyber Science, Nankai University,
  \textsuperscript{5}Hello Group\\
  \textsuperscript{6}School of Social Development and Public Policy, Fudan University\\
  \textsuperscript{$\heartsuit$}\texttt{weiyuanli25@m.fudan.edu.cn},
  \textsuperscript{$\spadesuit$}\texttt{yangdeqing@fudan.edu.cn}
}

\usepackage{xcolor}

\begin{document}
\maketitle
\begin{abstract}
Open-ended reinforcement learning often relies on rubric-based rewards for tasks without directly verifiable answers. Yet the policy and reward system form a dynamic feedback loop: as the policy optimizes the current reward, an initially useful reward system may become unreliable due to reward hacking or reduced response discriminability. The reward system should therefore evolve rather than remain fixed during training.
Existing dynamic-rubric methods adapt evaluation criteria, but reward failures can also arise from scoring mechanisms or signal composition. We introduce \method{}, a self-evolving RL framework that evolves the reward system from on-policy experience, representing it as an executable \rewarddag{}. Specifically, an agentic designer updates this system from on-policy rollouts and reward traces to maintain train-time reliability.
Across writing and roleplay, \method{} achieves the best quality
under all three judges, outperforming the policy by \(2.107\) and \(4.767\)
points, respectively, while reducing reward hacking and coverage failures and
preserving reward informativeness. Ablations confirm that a comprehensive
fixed reward system cannot remain reliable in open-ended tasks and must
evolve throughout training.

\end{abstract}

\section{Introduction}

\begin{figure}[t]
\includegraphics[
    width=\columnwidth,
    height=0.4\textheight
  ]{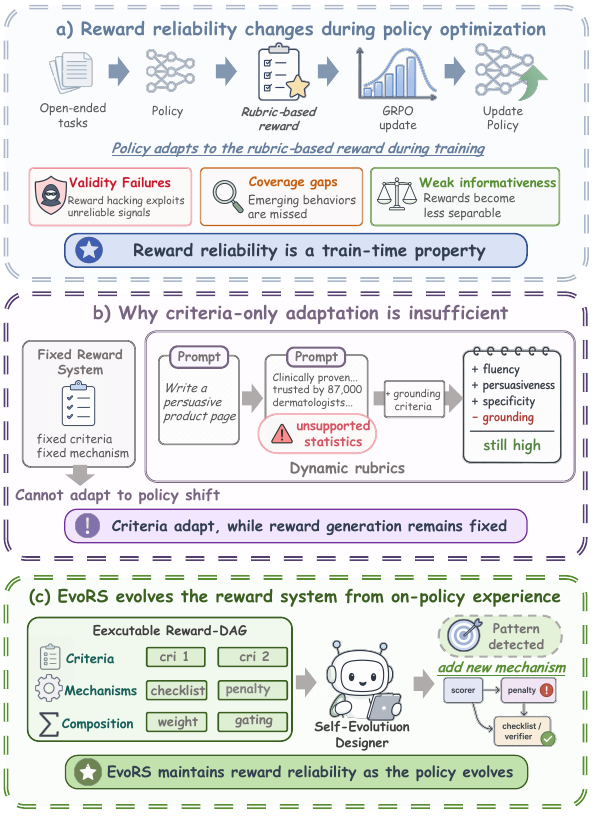}
  \caption{\textbf{Reward-system evolution in open-ended RL.} As policy learning exposes failures in reward validity, coverage, and informativeness, \method{} uses on-policy experience to generate and compare candidate reward-system states, allowing the reward system to evolve with the policy.}
  \label{fig:experiments_1}
\end{figure}

Reinforcement learning (RL) is increasingly used to train LLMs for open-ended tasks, from writing and roleplay to multi-step office workflows handled by work agents~\citep{jia2025writingzero,tseng2024two}.
Unlike mathematical reasoning or code generation, for which rewards can often be derived from verifiable answers, these tasks require judgments over multiple interacting dimensions that are difficult to specify completely. Existing methods therefore rely on rubric-based reward systems, using LLM judges or Generative Reward Models (GRMs) to turn such judgments into scalable training signals~\citep{zheng2023judging,gu2024survey,mahan2024generative,zhang2024genrm,gunjal2025rubrics}. However, rubric-based reward systems depend on how they are designed and may become less effective during training. This makes reward reliability a central concern in open-ended tasks.

During RL, the policy continually adapts to the current reward system. As the policy changes, a reward system that is reliable for the initial policy may become unreliable later in training. We analyze such failures along three dimensions. \textbf{Validity} fails when reward hacking lets the policy obtain high reward without producing the intended behavior~\citep{li2025curse,zhao2025onetoken}. \textbf{Coverage} becomes insufficient when emerging on-policy responses are not represented by the current reward system~\citep{lang2024fine}. \textbf{Informativeness} diminishes when the reward signal gradually becomes less useful for optimization, for example as rewards saturate or become less separable within same-query groups~\citep{kim2025rethinking}. Together, these failures show that reward reliability cannot be treated as fixed throughout training. The reward system should therefore evolve alongside the policy.

To address these training-time reliability failures, recent dynamic-rubric methods generate rubrics dynamically to cover response patterns missed by fixed rubrics~\citep{yu2025rewardanything,shao2025drtulu}. These methods mainly change the rubric criteria, while the scoring mechanisms that turn evaluations into rewards and the composition of multiple signals often remain fixed. Figure~\ref{fig:experiments_1} illustrates the gap: adding a grounding criterion for a response containing unsupported statistics can still leave its score high under continuous scoring, whereas a penalty mechanism can more selectively penalize such hacking patterns. Reward reliability therefore depends on the complete reward system, including its criteria, scoring mechanisms, and signal composition; these system-level choices should evolve during training rather than remain fixed.

We therefore introduce \method{}, a self-evolving reward-system framework for RL on open-ended tasks. \method{} treats reward design as part of the learning process rather than a one-time specification: the reward system evolves alongside the policy as training exposes new failures. To support this evolution, \method{} represents the complete reward system as an executable \rewarddag{}, allowing its criteria, scoring mechanisms, and signal composition to be revised within a unified structure.

To adapt the reward system as the policy changes, a reward-system designer analyzes on-policy rollouts and reward traces to diagnose weaknesses in the current state. Based on this diagnosis, it generates several targeted and bounded candidate \rewarddag{} states. Matched replay compares the current and candidate states on the same rollout cases, and the candidate that best repairs the targeted failure while preserving useful reward behavior becomes the next active state. If no candidate provides an improvement, the current state is retained. The resulting reward system supplies rewards for subsequent policy updates, allowing it to evolve alongside the policy. We instantiate \method{} on writing and roleplay, where it achieves the best final benchmark quality under all three judges. While prior methods exhibit increased reward hacking and uneven gains across roleplay dimensions, \method{} reduces reward-hacking and coverage-failure rates, preserves reward informativeness, and improves all four roleplay dimensions. Component-control ablations further confirm that these gains depend on evolving the reward system alongside the policy throughout training.

Our key contributions are as follows:
\begin{itemize}[leftmargin=*]
    \item We frame reward reliability in open-ended RL as a training-time problem: as the policy adapts to its reward, the reward system may suffer reward hacking, insufficient on-policy reward coverage, and diminishing reward informativeness.

    \item We introduce \method{}, a self-evolving reward-system framework that represents the complete reward system as an executable \rewarddag{} and searches for bounded candidate states from on-policy evidence. These candidates may revise criteria, scoring mechanisms, or signal composition, while comparative feedback determines the next reward-system state.

    \item We validate \method{} on writing and roleplay, showing that it
    improves final policy quality and reward reliability; controlled ablations
    further confirm the importance of evolving the reward system alongside the
    policy.
\end{itemize}

\section{Related Work}

\subsection{Rubric-Based Rewards and Dynamic Rubrics}
Rubric-based rewards use LLM judges or GRMs to provide scalable supervision for open-ended tasks, where quality is multi-dimensional and not directly verifiable. Prior work has instantiated this supervision through continuous rubric scores and principle-based judgments~\citep{arora2025healthbench,kim2024biggenbench,WritingBench,ref-and-rubrics}, checklist-based verification~\citep{judge:3,CheckEval,viswanathan2025rlcf}, penalty or constraint signals and hybrid designs~\citep{wang2025coser,RULERS,liu2025openrubrics}, and pairwise preference comparisons~\citep{xu2025pairwiseframework,jia2025writingzero,liu2025deepseekgrm,du2026her}. In RL, rubric-based methods have progressed from broad principles~\citep{bai2022constitutional,sun2023salmon,yu2025rewardanything} to task- or query-specific rubrics and anchors~\citep{gunjal2025rubrics,huang2025rubricanchors}. Recent methods generate or update rubrics dynamically using task information, response comparisons, or online policy outputs~\citep{rezaei2025onlinerubrics,lv2026queryspecific,xu2026rubricarm,jia2026openrs,shao2025drtulu,shen2026rethinking,ding2026evorubrics}. These methods make the rubric criteria more adaptive, while the scoring and composition procedures that convert them into training rewards generally remain unchanged. This line of work therefore leaves the broader construction of the reward signal outside its scope.

\subsection{Reward Reliability under Policy Optimization}
Reward reliability becomes more challenging when an evaluator is used as an RL objective. LLM judges and reward models exhibit systematic biases even under static evaluation~\citep{zheng2023judging,Bias-All:1,Bias-Position:1,wang2026rmbias}; under optimization, policies can exploit biases, loopholes, or underspecified objectives, leading to reward overoptimization or reward hacking~\citep{skalse2022rewardgaming,lambert2023alignmentceiling,gao2023scalinglaws,moskovitz2024confronting,hou2026prorl,mei2026good}. This risk is particularly relevant to rubric-based RL, where explicit criteria can expose shortcut patterns and fixed criteria can miss emerging on-policy responses~\citep{mahmoud2026rubrichacking,zhang2026chasingtaileffectiverubricbased,li2025curse,zhao2025onetoken,singhal2024longway,lang2024fine}. Moreover, reward-model accuracy on a fixed distribution does not ensure that the reward remains informative as the policy changes; useful RL signals require meaningful distinctions among on-policy responses, especially within the same query~\citep{levine2023distributionshift,chen2024accuracyparadox,razin2025goodteacher}. Together, these findings show that reward reliability must be evaluated on the responses generated during policy optimization, not only on fixed evaluation outputs.

\subsection{Self-Evolving Training and Agent Systems}
A broader line of self-evolving training work uses experience to adapt different parts of the learning loop. CoEvolve evolves training data, ECHO evolves a natural-language critic alongside the policy, and ReSkill evolves a skill bank within agentic RL~\citep{yang2026coevolve,li2026echo,he2026reskill}. Other work studies self-evolving harnesses that revise the runtime scaffolding or training environment around an agent; EvoTrainer extends this direction to training-side diagnostics and interventions across policy versions~\citep{chen2026evotrainer,zhang2026selfharness,huang2026envharness,chen2026traineetrainer}. Across these settings, experience is used to identify what the current training setup lacks and guide targeted, validated changes. Our work applies this principle to the reward system itself: the complete system that supplies the policy's training signal during RL.

\section{Motivating Analysis}

The central challenge is to keep the reward system reliable as the policy adapts to it. This requires understanding both why a fixed reward system becomes insufficient during training and how on-policy experience can guide its evolution.

\begin{figure}[t]
  \includegraphics[width=\columnwidth]{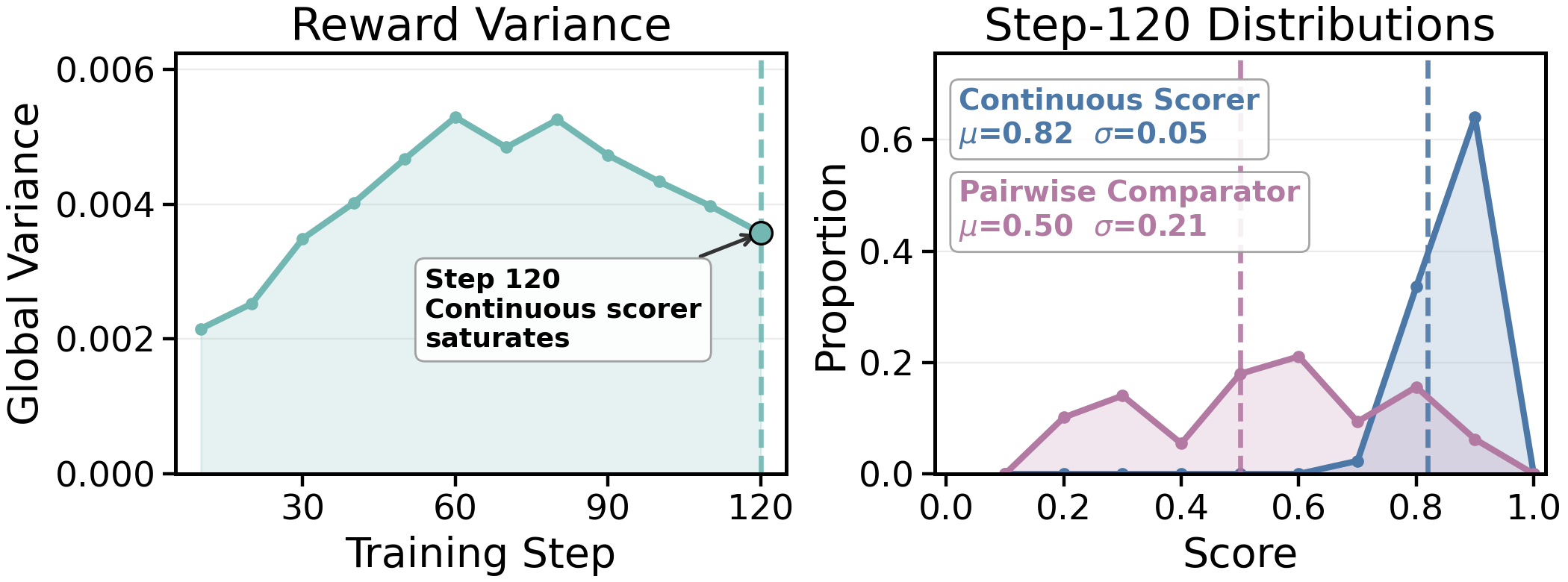}
  \caption{\textbf{Fixed scorers can saturate.} Later responses concentrate in a narrower score region, reducing reward variance and sample-level separability.}
  \label{fig:experiments_3}
\end{figure}

\paragraph{The need for reward-system evolution.}
During RL, the current reward system shapes each policy update, while the updated policy changes the response distribution on which that reward system operates. Figure~\ref{fig:experiments_3} illustrates this change: a reward system that provides useful discriminability earlier in training may later fail to maintain reward variance and sample-level separability, making it difficult to provide an effective reward signal. Thus, the reward system is not a fixed component that can be determined once before training; it must continue to evolve with the policy.

\begin{figure}[t]
  \includegraphics[width=\columnwidth]{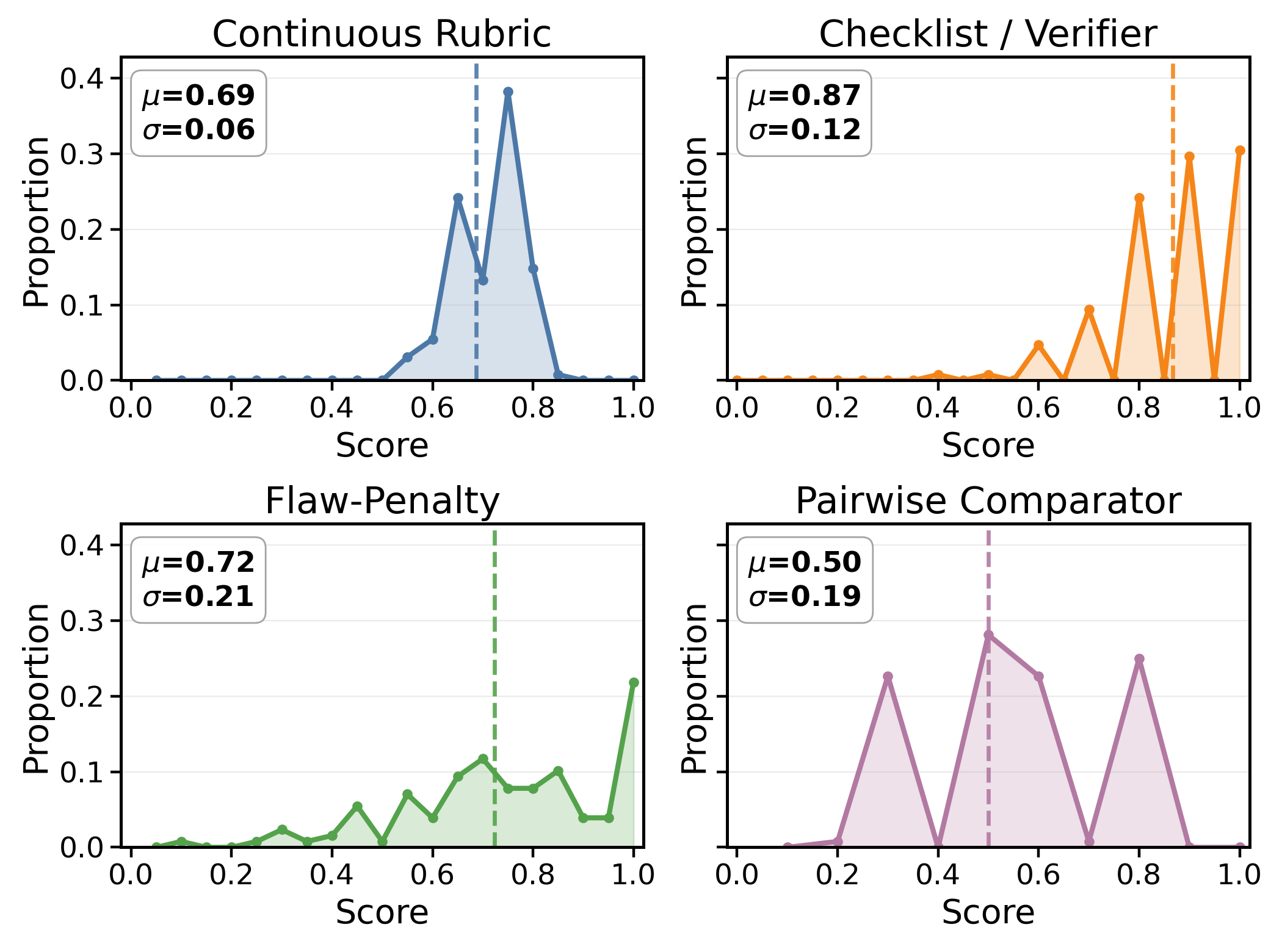}
  \caption{\textbf{Mechanisms are not interchangeable.} Different mechanisms induce different score distributions on the same on-policy responses, affecting density, same-query discriminability, and failure sensitivity.}
  \label{fig:experiments_2}
\end{figure}

In rubric-based reward systems for open-ended tasks, a rubric describes what a good response should satisfy, but it does not by itself determine the reward that the policy receives. The same evaluation content can be turned into a reward in different ways: a judge can assign continuous scores, check requirements one by one or deduct points for specific flaws. Table~\ref{tab:reward_mechanisms} summarizes these reward mechanisms, and Figure~\ref{fig:experiments_2} shows that they produce different reward distributions on the same responses. In practice, the mechanisms can also complement one another. For example, a checklist can ensure that necessary requirements are met, while continuous scoring distinguishes quality among responses that pass those checks. Reward reliability therefore depends on the whole path from what is evaluated, to how it is scored, to how multiple signals are combined. What should evolve is not only the rubric or a single scorer, but the complete reward system. Appendix~\ref{app:additional_motivating_results} provides additional analyses of mechanism-specific separability and zero-distinction patterns.

\begin{table}[t]
\centering
\small
\setlength{\tabcolsep}{0pt}
\begin{tabular*}{\columnwidth}{@{\extracolsep{\fill}}lll@{}}
\toprule
Mechanism & Signal & Typical Risk \\
\midrule
Continuous Scorer & Dense score & Saturation \\
Checklist Verifier & Pass rate & Sparsity\\
Flaw-Penalty & Failure penalty & Over-penalty \\
Pairwise Comparator & Preference signal & Pair dependence \\
\bottomrule
\end{tabular*}
\caption{\textbf{Reward mechanisms expose different signals.} We consider dense scorer saturation, checklist sparsity, flaw-penalty over-penalization, and pair dependence.}
\label{tab:reward_mechanisms}
\end{table}

\paragraph{From failure evidence to a targeted repair.}
Observing a reward failure does not by itself reveal how the reward system should change. For example, low within-query variance may arise because the criteria are too generic to express quality differences among responses, or because scorer saturation compresses different responses into similar scores. The former calls for refining the criteria, whereas the latter calls for changing the scoring mechanism. Reward-system evolution must therefore be based on diagnosis of on-policy rollouts and their reward traces, rather than mapping a single failure indicator directly to an edit. When the evidence does not support a clear diagnosis, the current reward system remains unchanged.

Diagnosis can identify a possible direction for changing the reward system, but it does not establish that the change will improve the reward. Reward-system evolution must also learn from the observed effects of candidate updates. The current and candidate reward systems are evaluated on the same rollout cases to determine whether the candidate repairs the targeted failure while preserving reward behavior that remains useful. This feedback determines which reward-system state is carried into subsequent training and provides experience for later updates. By repeatedly proposing, evaluating, and selecting candidate updates, the reward system can continue to evolve with the policy. Together, diagnosis and comparative validation turn on-policy experience into iterative reward-system evolution. We instantiate this process in \method{}, described next.

\begin{figure*}[!t]
  \centering
  \includegraphics[
    width=1\textwidth,
    height=0.34\textheight
  ]{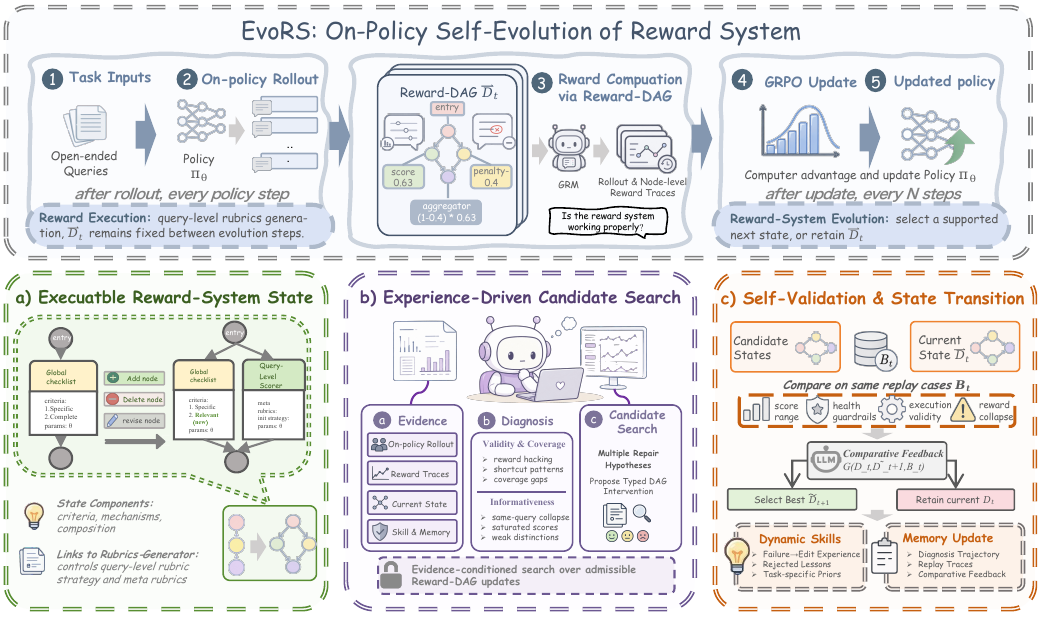}
  \caption{
  \textbf{\method{} overview.} The top panel couples policy learning with reward-system evolution. The bottom panels show how \method{} represents the current reward-system state as an executable \rewarddag{}, generates alternative candidate states from on-policy experience, and uses comparative feedback to select the next state while accumulating memory and dynamic skills for future search.
  }
\label{fig:dream_framework}
\end{figure*}

\section{Method}

We introduce \method{}, a framework that couples policy learning with reward-system evolution. During RL, the current policy \(\pi_t\) generates groups of responses, which the current reward-system state \(D_t\) evaluates to produce both training rewards and node-level reward traces. These responses and traces form the on-policy experience \(\mathcal O_t\), revealing how the reward system behaves on the policy distribution it currently shapes.

As training proceeds, \method{} revisits the reward system every \(N\) policy updates and uses this experience to determine whether it should evolve. When an update is needed, it diagnoses the observed evidence, proposes alternative candidate states, and compares their effects through matched replay to select the next reward-system state \(D_{t+1}\). Otherwise, the current state \(D_t\) continues to serve the policy. Figure~\ref{fig:dream_framework} illustrates this closed loop.

\subsection{The Reward System as an Evolving State}
\label{sec:method_reward_dag}

To make the complete reward system evolvable with the policy, we abstract its current state as an executable \rewarddag{}:

\begin{equation}
D_t=(\mathcal{V}_t,\mathcal{E}_t,\Theta_t),
\end{equation}

where \(\mathcal{V}_t\) contains executable reward nodes, \(\mathcal{E}_t\) specifies their dependencies and execution order, and \(\Theta_t\) defines the criteria and scoring mechanism used by each node, together with the rules for combining node outputs. Thus, \(D_t\) represents in one state what is evaluated, how each reward signal is produced, and how these signals form the final reward. By representing the complete reward system as a \rewarddag{}, \method{} turns the reward system itself into an evolvable object.

A Reward-DAG contains Rubric Nodes that produce reward signals and Composition Operators that combine them. Each Rubric Node \(v\in\mathcal{V}_t\) instantiates criteria \(\rho_v\) and applies a scoring mechanism \(m_v\) to produce a node-level signal:
\begin{equation}
h_v=M_{m_v}(x,y;\rho_v,\theta_v,J_\phi).
\end{equation}

Here, \(\rho_v\) may be fixed at the task level or instantiated by \(G_\psi(x)\) for the current query. Each node evaluates one response at a time; response groups are used by GRPO and by the diagnostic and comparison procedures described below. The outputs and execution records of all nodes constitute the node-level reward trace introduced above.

A Composition Operator maps upstream node outputs to the scalar reward used by RL:
\begin{equation}
r_D(x,y)=F_{\mathrm{out}}(\{h_v:v\in \mathrm{Pa}(\mathrm{out})\};\theta_{\mathrm{out}}).
\end{equation}

It does not evaluate the response again, but determines how existing signals are aggregated, normalized, gated, or routed. Because criteria, scoring mechanisms, and signal combination are represented in the same Reward-DAG, reward-system evolution can modify any of them within a unified representation.

Crucially, Reward-DAG separates reward execution from agentic evolution. The agentic designer uses accumulated on-policy experience only to propose candidate states and does not participate in reward computation itself. Between evolution steps, the active \(D_t\) remains fixed. By keeping reward execution independent of the designer, reward computation remains consistent within each training stage and comparable across candidate states. Appendix~\ref{app:reward_dag} provides the execution schema and node-level details; the next section describes how on-policy experience gives rise to candidate reward-system states.

\subsection{From On-Policy Experience to Candidate States}
\label{sec:method_meta_judge_policy}

To maintain reward-system reliability during training, \method{} uses an agentic
designer to turn accumulated on-policy experience \(\mathcal O_t\) into a
structured diagnosis and a potentially useful revision direction for the current
reward-system state \(D_t\). By jointly analyzing sampled responses and their
reward traces, the designer relates recurring response patterns to how \(D_t\)
produces the training signal,
then identifies the primary reward failure and its likely source in the
Reward-DAG. If the evidence does not support a targeted update, \(D_t\) remains
unchanged.

Conditioned on the diagnosis, the designer explores alternative repair
directions within the admissible update space \(\mathcal A(D_t)\). It instantiates
each direction as a typed update \(e_t^{(j)}\) and the corresponding
candidate state \(\widetilde D_{t+1}^{(j)}\):
\begin{equation}
\begin{aligned}
\{e_t^{(j)}\}_{j=1}^{M}
&=\Pi_{\mathrm{design}}(D_t,\mathcal O_t,\mathcal K_t),\\
\widetilde D_{t+1}^{(j)}&=e_t^{(j)}(D_t),
\qquad e_t^{(j)}\in\mathcal A(D_t).
\end{aligned}
\end{equation}
Here, \(\mathcal K_t\) comprises the designer's run-local memory and dynamic
skills accumulated from previous evolution steps. Each candidate represents a
targeted change to the reward system, but remains a hypothesis until its actual
reward behavior is observed. The next section describes how comparative
self-validation selects the next reward-system state.

\subsection{Self-Validation and Evolutionary State Transition}
\label{sec:method_replay_evolution}

To turn these hypotheses into an evolution step, \method{} evaluates alternative
reward-system states through the reward behavior they actually produce. It
applies the current state \(D_t\) and all candidate states
\(\{\widetilde D_{t+1}^{(j)}\}_{j=1}^{M}\) to the same replay cases
\(\mathcal B_t\) drawn from recent on-policy experience. Holding the rollout
cases fixed makes their effects directly comparable, revealing whether a
candidate addresses the diagnosed failure and how it changes the reward
behavior of the current system.

Based on this comparative feedback, \method{} selects as \(D_{t+1}\) the
candidate that most consistently corrects the targeted failure cases while
preserving rewards on clean cases and the within-query distinctions needed for
group-relative policy learning. If no candidate provides a supported
improvement, the current state is retained. Each selected, rejected, or no-op
trajectory is stored in run-local memory and distilled into dynamic evolution
skills, which together form \(\mathcal K_{t+1}\). By carrying this experience
into subsequent diagnoses, the designer progressively adapts how it searches
for reward-system improvements. Matched replay therefore provides feedback not
only for selecting \(D_{t+1}\), but also for improving the process that
generates future candidate states.
Appendix~\ref{app:reward_dag} specifies the replay construction and comparison
criteria; Appendix~\ref{app:algorithm} gives the full evolution loop.

\section{Experiments and Results}
\label{sec:experiments}
\subsection{Experimental Setup}
\label{sec:exp_setup}

\begin{table*}[!t]
\centering
\small
\setlength{\tabcolsep}{3.2pt}
\resizebox{\textwidth}{!}{%
\begin{tabular}{lcccccccccccc}
\toprule
& \multicolumn{4}{c}{Writing} & \multicolumn{8}{c}{Roleplay (CoSER)} \\
\cmidrule(lr){2-5}\cmidrule(lr){6-13}
Method
& \multicolumn{1}{c}{Quality}
& \multicolumn{2}{c}{Reliability}
& \multicolumn{1}{c}{Signal}
& \multicolumn{5}{c}{Quality}
& \multicolumn{2}{c}{Reliability}
& \multicolumn{1}{c}{Signal} \\
\cmidrule(lr){2-2}\cmidrule(lr){3-4}\cmidrule(lr){5-5}\cmidrule(lr){6-10}\cmidrule(lr){11-12}\cmidrule(lr){13-13}
& WB & HR & CFR & R-Var & Overall & SC & AN & CF & SQ & HR & CFR & R-Var \\
\midrule
\multicolumn{13}{l}{\textit{Base model}} \\
Qwen3-4B base
& 54.894 & 7.7 & 27.8 & --
& 60.909 & 59.097 & 64.172 & 60.487 & 59.881 & 6.06 & 13.0 & -- \\
\midrule
\multicolumn{13}{l}{\textit{Static rubrics}} \\
RLAIF
& 56.276 & 13.8 & \underline{21.8} & 0.0010
& 59.906 & 59.065 & 61.889 & 61.704 & \underline{56.966} & \underline{7.43} & 16.0 & 0.0029 \\
RaR
& \underline{56.547} & 14.7 & 29.5 & 0.0014
& \underline{60.000} & 60.704 & 61.752 & 62.478 & 55.068 & 8.52 & 22.0 & 0.0061 \\
\midrule
\multicolumn{13}{l}{\textit{Dynamic rubrics}} \\
RLER
& 56.170 & 11.0 & 23.9 & 0.0011
& 59.621 & \underline{60.868} & 59.779 & \underline{63.364} & 54.474 & 9.58 & 18.0 & 0.0137 \\
OpenRS
& 56.427 & \underline{10.1} & 31.1 & --
& 58.598 & 57.456 & \underline{62.614} & 59.334 & 54.986 & 8.83 & \underline{15.5} & -- \\
\midrule
\multicolumn{13}{l}{\textit{Our method}} \\
\textbf{\method{}}
& \textbf{57.001} & \textbf{6.5} & \textbf{19.9} & 0.0025
& \textbf{65.676} & \textbf{63.615} & \textbf{69.880} & \textbf{63.381} & \textbf{65.829} & \textbf{5.25} & \textbf{10.0} & 0.0063\\
\bottomrule
\end{tabular}%
}
\caption{\textbf{Overall policy quality and reward reliability.} We evaluate
final policy quality on WritingBench (WB) and CoSER using GPT-5.6-Terra,
DeepSeek-V4-Pro, and GLM-5.2 as judges and average their scores; CoSER Overall
aggregates its four roleplay dimensions. HR and CFR assess reward validity and
coverage, respectively, while R-Var measures within-query reward separation.}
\label{tab:main_results}
\end{table*}

\paragraph{Tasks, data, and evaluation.}
We evaluate \method{} on two open-ended domains: writing and roleplay. Writing
uses 944 filtered synthetic instructions, while roleplay uses 5,000 queries
from MiniMax Role-Play Bench; Appendix~\ref{app:data_and_rubrics} provides the
full data construction and distribution. We evaluate final policy quality using
WritingBench~\citep{WritingBench} and CoSER~\citep{wang2025coser}, with multiple
strong judge models on both tasks to reduce dependence on any single evaluator.
We report the overall WritingBench score, together with CoSER Overall and its
SC, AN, CF, and SQ dimensions. The complete evaluation protocol and judge
configurations are provided in
Appendix~\ref{app:benchmark_score_uncertainty}.

\paragraph{Training and reward systems.}
We compare \method{} with RLAIF~\citep{bai2022constitutional},
RaR~\citep{gunjal2025rubrics}, RLER~\citep{shao2025drtulu}, and
OpenRS~\citep{jia2026openrs}. RLAIF uses a fixed task-level rubric, while RaR
uses fixed query-specific rubrics. RLER augments
each task rubric with additional criteria elicited from current policy
responses, whereas OpenRS constructs pairwise adaptive rubrics from response
comparisons. All methods use the same GRPO pipeline, training data, rollout
budget, and evaluation protocol. They use the same local Qwen3.5-27B model as
the GRM for rubric-based reward computation. The reward system of \method{}
starts from the same single-node continuous reward system as RLAIF and evolves
every five RL steps. The appendices provide the full training-data distribution
and experimental configurations
(Appendices~\ref{app:data_and_rubrics} and~\ref{app:implementation_details}).

\paragraph{Metrics and audits.}
We use WritingBench and CoSER to evaluate final response quality. Following
prior analyses and designs of exploit rates and reward variance, we evaluate
reward reliability from three dimensions: HR measures the occurrence of
predefined reward-hacking patterns~\citep{mahmoud2026rubrichacking,thaman2026rewardhackingbenchmark},
CFR measures predefined hard failures that the reward may
under-cover~\citep{zhou2023ifeval,jiang2023followbench}, and R-Var measures
within-query reward separation~\citep{razin2025goodteacher,xu2025notallrollouts}.
The HR/CFR categories are constructed before method comparison and applied
blindly to all outputs. Appendix~\ref{app:hack_aware_audit} provides details.

\subsection{Main Results}
\label{sec:exp_main_results}

Table~\ref{tab:main_results} compares different methods in terms of final policy
quality and reward reliability. On WritingBench, all RL methods improve over the
base policy, while \method{} produces the largest quality gain of \(2.107\)
points and is the only method that reduces HR below the base level. The contrast
is sharper on CoSER. Methods trained with fixed or partially adapted reward
systems do not improve overall quality over the base policy and exhibit higher
HR. Their improvements are also uneven across dimensions: for example, RaR
improves SC and CF but reduces AN and SQ. This pattern is consistent with
overoptimization of an increasingly exploitable reward signal---the policy
learns the aspects emphasized by the current reward while sacrificing behaviors
it fails to capture. In contrast, \method{} improves CoSER Overall by \(4.767\)
points over the base policy and ranks first across all four dimensions.

These results demonstrate the benefit of \textbf{maintaining a reward system
that remains effective for the evolving policy rather than continuing to
optimize fixed reward criteria}. Appendix~\ref{app:benchmark_score_uncertainty}
provides judge-specific results, paired uncertainty estimates, and further
analysis of the roleplay outputs. The next section examines how reward
reliability changes throughout training.

\section{Analysis}
\label{sec:analysis}

\subsection{Reward-System Evolution under Policy Shift}
\label{sec:exp_dynamics}

\begin{figure}[t]
\centering
\includegraphics[width=\columnwidth]{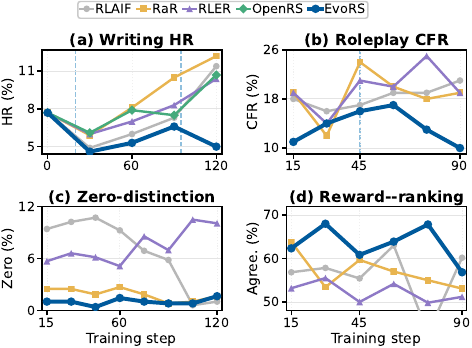}
\caption{\textbf{Training-time reward dynamics.} (a,b) \method reduces writing
HR and roleplay CFR after selected reward-system transitions (dashed lines). (c,d) Its
rewards preserve same-query distinctions and stronger agreement with blind
preference rankings.}
\label{fig:training_time_dynamics}
\end{figure}

Figure~\ref{fig:training_time_dynamics} traces reward validity, coverage, and
informativeness as the policy changes, allowing us to examine whether newly
exposed reward failures are followed by corresponding changes in the
reward-system state.

\paragraph{Validity and coverage.}
In writing, reward-hacking patterns increase under both static and
dynamic-rubric baselines, whereas \method{} remains lower and declines further
after the annotated reward-system transition. In roleplay, \method{} similarly
maintains a lower CFR and shows a further reduction after the step-45 coverage
repair. Because the audit categories are hidden from the designer, these
trajectories reflect failures discovered from on-policy experience rather than
direct optimization of the reported metrics.
Appendix~\ref{app:case_studies} provides detailed examples of these diagnoses
and state changes.

\paragraph{Informativeness.}
\label{sec:exp_signal_quality}
Figure~\ref{fig:training_time_dynamics}(c,d) complements the final R-Var results
by examining whether within-query distinctions remain both available and
meaningful during training. Across the 480-group audit, \method{} maintains a
low Zero-distinction rate and stronger agreement with blind same-query
preference rankings across most checkpoints. The evolved reward therefore does
not merely produce more dispersed scores; its distinctions remain aligned with
independent response preferences.
Appendix~\ref{app:reward_signal_ranking_audit} provides the evaluation protocol
and reliability checks.

\subsection{Generality across Policy and Reward Models}
\label{sec:exp_generality}

We further examine whether the benefit of reward-system evolution depends on
the Qwen3-4B policy or the Qwen3.5-27B reward model used in the main
experiments. We scale the policy to Qwen3-8B~\citep{yang2025qwen3} and evaluate
both Qwen3.5-27B and RewardAnything-8B~\citep{yu2025rewardanything} as GRMs. As
shown in Table~\ref{tab:generality}, \method{} outperforms RaR under both reward
models and also improves over the Qwen3-8B base policy. The advantage of
reward-system evolution therefore persists across policy scales and GRM
families.

\begin{table}[t]
\centering
\small
\setlength{\tabcolsep}{4.0pt}
\begin{tabular}{llr}
\toprule
Method & GRM & CoSER Overall \\
\midrule
Qwen3-8B base & -- & 64.545 \\
\midrule
RaR & Qwen3.5-27B & 62.044 \\
\textbf{\method{}} & Qwen3.5-27B & \textbf{66.370} \\
\midrule
RaR & RewardAnything-8B & 62.761 \\
\textbf{\method{}} & RewardAnything-8B & \textbf{65.660} \\
\bottomrule
\end{tabular}
\caption{\textbf{Generality across policy and reward models.} CoSER Overall is
averaged across the same three judges as Table~\ref{tab:main_results}.}
\label{tab:generality}
\end{table}

\subsection{Ablation Studies}
\label{sec:exp_ablation}

We ablate both the scope of reward-system evolution and the capabilities that
support its search over improved reward-system states.

\paragraph{Evolution strategy.}
\label{sec:exp_ablation_downstream}

We first examine what should evolve and when this evolution is needed. Only
criteria-level adaptation retains the same evolution process but restricts its
search space to rubric criteria, while Fixed final \rewarddag{} uses the
complete reward system discovered by \method{} from the beginning of training.
Both strategies underperform full reward-system evolution, with particularly
large drops on CoSER. A comprehensive reward system fixed in advance therefore
cannot replace the evolution path, and adapting evaluation content alone cannot
address failures arising from scoring mechanisms or signal composition.

\begin{table}[t]
\centering
\small
\setlength{\tabcolsep}{3.5pt}
\resizebox{\columnwidth}{!}{%
\begin{tabular}{lrrrr}
\toprule
Variant & WB & \(\Delta\) & CoSER & \(\Delta\) \\
\midrule
Full \method{} & \textbf{57.001} & -- & \textbf{65.676} & -- \\
\midrule
\multicolumn{5}{l}{\textit{Evolution strategy}} \\
only criteria-level adaptation & 56.593 & \(-0.408\) & 61.155 & \(-4.521\) \\
Fixed final \rewarddag{} & 56.308 & \(-0.693\) & 60.573 & \(-5.103\) \\
\midrule
\multicolumn{5}{l}{\textit{Self-evolution components}} \\
w/o candidate selection & 56.959 & \(-0.042\) & 62.494 & \(-3.182\) \\
w/o update rejection & 56.807 & \(-0.194\) & 63.911 & \(-1.765\) \\
w/o evolution memory / skills & 56.918 & \(-0.083\) & 62.412 & \(-3.264\) \\
\bottomrule
\end{tabular}%
}
\caption{\textbf{Downstream ablations.} WB and CoSER report three-judge
average scores, and \(\Delta\) is relative to the full method. Detailed CoSER
dimensions and reward-reliability audits are reported in
Appendix~\ref{app:ablation_details}.}
\label{tab:downstream_component_ablation}
\end{table}

\paragraph{Self-evolution components.}
We then examine how \method{} searches for its next reward-system state.
Candidate selection compares alternative repair directions through their
observed reward behavior; update rejection preserves the current state when no
candidate provides an improvement; and evolution memory makes previous outcomes
available to later search. Removing any of these capabilities reduces final
quality, with substantially larger effects on multi-turn roleplay. The
dimension-level results in Appendix~\ref{app:ablation_details} further show that
some variants retain high SC or CF while sharply reducing AN and SQ, indicating
less balanced optimization.

\section{Conclusion}

We presented \method{}, a framework for on-policy self-evolution of reward
systems in open-ended RL. It represents the active reward system as an
executable \rewarddag{}, diagnoses its failures from on-policy reward traces,
and searches for bounded candidate states. Comparative feedback on matched
rollout cases then determines the next active state, closing the loop between
policy learning and reward-system evolution.

Our experiments reveal a broader lesson: even a comprehensive reward system
specified before training underperforms one that evolves with the policy. This
also explains the limitation of recent dynamic-rubric methods: although they
adapt what is evaluated, they leave how reward signals are produced and
composed unchanged, resulting in uneven quality gains. \method{} realizes a
broader self-evolution paradigm by diagnosing failures from on-policy reward
traces and searching for improvements across the full reward-system state.

\section*{Limitations}

Our experiments establish reward-system self-evolution in open-ended language
generation, while leaving several extensions for future work.

The current evaluation focuses on writing and roleplay. Beyond these domains,
the executable \rewarddag{} can in principle incorporate tool state,
interaction history, and delayed outcomes, giving \method{} the potential to
extend to agentic RL settings. This potential still requires validation in
tool-using environments.

In our experiments, reward-system evolution is performed every \(N\) RL steps
to simplify the system design and control computational cost. A schedule
conditioned on accumulated failure evidence could instead adapt the timing of
evolution to the changing policy, which we leave for future study.

Periodic diagnosis and candidate evaluation also introduce some additional
training cost. Appendix~\ref{app:runtime_cost} reports the wall-clock
performance comparison with other methods. Adaptive scheduling or batched
candidate evaluation may further reduce this overhead.


\bibliography{reference}

\clearpage
\appendix

\section{Appendix Roadmap and Robustness Checks}
\label{app:validity_roadmap}

Because the evaluation involves open-ended RL, LLM judges, and a self-evolving
reward-system loop, we include additional checks for evaluation robustness,
auditor blinding, reward-evolution traceability, baseline instantiation, and
reproducibility. Table~\ref{tab:validity_roadmap} gives a roadmap to these
materials. It is intended as a navigation aid rather than as additional
experimental evidence.

\begin{table*}[!tbp]
\centering
\small
\setlength{\tabcolsep}{3.0pt}
\resizebox{\textwidth}{!}{%
\begin{tabular}{p{0.18\textwidth}p{0.38\textwidth}p{0.29\textwidth}p{0.10\textwidth}}
\toprule
Appendix topic & What is covered & Purpose & Details \\
\midrule
Evaluation Protocols
& Multi-judge WritingBench and CoSER protocols, common-success aggregation, judge-specific results, paired uncertainty estimates, roleplay output analysis, and blinded HR/CFR audits.
& Documents how benchmark quality, robustness diagnostics, judge variance, and audit blinding are handled.
& App.~\ref{app:hack_aware_audit}; App.~\ref{app:benchmark_score_uncertainty} \\
\midrule
Training-Time Evidence
& Selected and non-selected candidate states, reward-variance dynamics, and blind reward-ranking audits over same-query rollout groups.
& Makes reward-system state transitions traceable and separates final-output gains from process-level evolution evidence.
& App.~\ref{app:candidate_state_trace_analysis}; App.~\ref{app:reward_signal_ranking_audit} \\
\midrule
Implementation and Reproducibility
& Model roles, training settings, prompts, Reward-DAG schema, typed edit operations, legality checks, candidate-comparison thresholds, runtime cost, and designer tools/skills.
& Specifies the reward-system interface and constrained designer workflow used in the experiments.
& App.~\ref{app:implementation_details}; App.~\ref{app:reward_dag}; App.~\ref{app:tools}; App.~\ref{app:skills} \\
\midrule
Baselines and Task Setup
& Artifact use, data safety, writing and roleplay training data, initial rubric construction, unified GRPO setup, and reward-system instantiations of static and dynamic-rubric baselines.
& Clarifies the shared experimental setting and how baseline reward systems are instantiated within it.
& App.~\ref{app:artifact_use}; App.~\ref{app:data_and_rubrics}; App.~\ref{app:baseline_details} \\
\midrule
Case Studies
& Writing guard-style penalty repair and roleplay coverage repair, including the diagnosed rollout evidence and selected Reward-DAG states.
& Illustrates how \method{} turns on-policy reward failures into targeted candidate states.
& App.~\ref{app:case_studies} \\
\bottomrule
\end{tabular}
}
\caption{Roadmap of appendix materials for evaluation protocols, training-time evidence, implementation details, baselines, and case studies.}
\label{tab:validity_roadmap}
\end{table*}

\section{Supplementary Motivating Analysis}
\label{app:motivating_analysis}

\subsection{Reward Mechanism Taxonomy}
\label{app:reward_mechanism_taxonomy}

This subsection clarifies the reward-mechanism space used in the motivating analysis. We use \textit{reward mechanism} to denote the scoring or judgment procedure that turns criteria and response evidence into an optimization signal. A mechanism specifies the elicited judgment, the granularity of the produced signal, and its normalization into a reward or preference signal. This is distinct from the \textit{criterion source}, which determines what evidence or criteria are evaluated, and from the \textit{composition operator}, which determines how multiple local signals are aggregated, gated, or routed.

This distinction is important for interpreting dynamic-rubric methods. Query-level rubrics, retrieved references, and generated checklists change the criterion source. Weighted sums, product guards, and routing modules change composition. A reward mechanism changes how judgments over criteria are elicited and exposed to the policy. Thus two reward systems can evaluate similar criteria but induce different training signals, for example by using dense scalar scores, pass-rate vectors, penalty scores, pairwise preferences, listwise ranks, or claim-level support vectors.

Table~\ref{tab:reward_mechanism_taxonomy} summarizes the mechanisms considered in this work and representative related work. The table is intended as a taxonomy of common scoring protocols rather than a claim that all mechanisms are used as live RL rewards in our experiments.

\begin{table*}[!tbp]
\centering
\small
\setlength{\tabcolsep}{3pt}
\begin{tabular}{p{0.17\textwidth}p{0.20\textwidth}p{0.16\textwidth}p{0.22\textwidth}p{0.17\textwidth}}
\toprule
Mechanism & Output signal & Typical granularity & Best suited for & Representative references \\
\midrule
Continuous / pointwise rubric scorer
& Dense scalar or dimension-wise scalar
& Response or rubric-dimension level
& Broad open-ended quality, writing quality, general rubric scoring
& BiGGen Bench, WritingBench, Prometheus, DeepSeek-GRM~\citep{kim2024biggenbench,WritingBench,ref-and-rubrics,liu2025deepseekgrm} \\
\midrule
Checklist / verifier scorer
& Pass/fail vector, pass rate, or partial-credit pass rate
& Checklist-item or criterion level
& Explicit requirements, constraint satisfaction, coverage of necessary elements
& RocketEval, CheckEval, RLCF, HealthBench, Rubrics as Rewards~\citep{judge:3,CheckEval,viswanathan2025rlcf,arora2025healthbench,gunjal2025rubrics} \\
\midrule
Flaw-penalty / deduction scorer
& Downward penalty score with violation trace
& Flaw, defect, or violation level
& Sparse bad-tail failures, roleplay violations, reward-hacking artifacts
& CoSER for penalty-based roleplay judging; reward-hacking work as motivation~\citep{wang2025coser,mahmoud2026rubrichacking,zhang2026chasingtaileffectiverubricbased} \\
\midrule
Pairwise preference comparator
& Pairwise winner, preference label, or win-rate reward
& Response-pair level
& Same-query relative quality and subtle writing or roleplay differences
& MT-Bench / Chatbot Arena, pairwise RLHF, Writing-Zero, HER, LitBench~\citep{zheng2023judging,xu2025pairwiseframework,jia2025writingzero,du2026her,LitBench} \\
\midrule
Listwise / ranking scorer
& Full ranking, top-\(k\), or rank-normalized score
& Response-set level
& Ordering a set of candidate responses under the same query
& LINKAGE and permutation-consensus listwise judging~\citep{LINKAGE,PCFJudge} \\
\midrule
Claim-level factuality / grounding scorer
& Claim-support vector, factual precision, unsupported-claim rate, or groundedness score
& Atomic-claim, QA-probe, citation, or evidence-unit level
& Long-form factuality, summarization consistency, RAG, deep research, citation grounding
& FActScore, SAFE/LongFact, QAGS, \(Q^2\), SummaC, DeepResearch Bench~\citep{FActScore,SAFE,QAGS,Q2,SummaC,DeepResearchBench} \\
\bottomrule
\end{tabular}
\caption{Representative reward mechanisms for open-ended and partially grounded generation. The taxonomy is organized by scoring protocol and output signal, not by whether a paper uses the word ``rubric'' or ``reward''. Criterion-source adaptation methods are discussed separately from reward mechanisms.}
\label{tab:reward_mechanism_taxonomy}
\end{table*}

\paragraph{Continuous / pointwise rubric scorer.}
The most common rubric-based protocol asks a judge to assign scalar scores to a response, either globally or along multiple rubric dimensions. With criteria \(\{c_k\}_{k=1}^{K}\), a typical normalized score is
\[
h_{\mathrm{cont}}(x,y)=\frac{1}{K}\sum_{k=1}^{K}s_k,\qquad s_k\in[0,1].
\]
This dense signal is easy to scalarize for RL and covers broad quality dimensions, but it can saturate as the policy improves, reducing same-query separability. Fine-grained LLM evaluation benchmarks and reward models commonly instantiate this mechanism through dimension-wise or rubric-conditioned scalar scores~\citep{kim2024biggenbench,WritingBench,ref-and-rubrics,liu2025deepseekgrm}. Evidence-anchored robust scoring systems such as RULERS are also related, but they should be viewed as robust pointwise scoring rather than penalty mechanisms~\citep{RULERS}.

\paragraph{Checklist / verifier scorer.}
Checklist mechanisms decompose the evaluation into explicit requirements and expose item-level satisfaction:
\[
h_{\mathrm{check}}(x,y)=\frac{1}{K}\sum_{k=1}^{K} z_k,\qquad z_k\in\{0,1\},
\]
with graded variants allowing \(z_k\in\{0,0.5,1\}\). This produces an interpretable pass vector or pass rate and is useful when necessary conditions must be checked explicitly. RocketEval and CheckEval study checklist-based LLM evaluation, RLCF uses checklist-style rewards for alignment, and HealthBench uses physician-written criteria that are judged as met or unmet before aggregation~\citep{judge:3,CheckEval,viswanathan2025rlcf,arora2025healthbench}. Rubrics as Rewards provides an RL example of structured rubric rewards beyond verifiable domains~\citep{gunjal2025rubrics}.

\paragraph{Flaw-penalty / deduction scorer.}
Deduction mechanisms start from an implicit or explicit full score and subtract penalties for detected flaws:
\[
h_{\mathrm{pen}}(x,y)=\mathrm{clip}\!\left(1-\lambda\sum_j \alpha_j v_j,0,1\right),
\]
where \(v_j\) denotes violation severity and \(\alpha_j\) denotes the corresponding weight. Unlike checklist scoring, the output is a downward correction targeted at sparse failures. CoSER is a direct example in roleplay evaluation: it identifies roleplay mismatch cases such as role-setting, language-style, action, or psychological mismatches and uses penalty-based LLM judging~\citep{wang2025coser}. Reward-hacking analyses motivate such bad-tail mechanisms, although they are not themselves deduction scorers~\citep{mahmoud2026rubrichacking,zhang2026chasingtaileffectiverubricbased}.

\paragraph{Pairwise preference comparator.}
Pairwise mechanisms elicit relative judgments between two responses:
\[
\begin{aligned}
p_{ij} &= J_\phi(x,y_i,y_j),\\
h_{\mathrm{pair}}(x,y_i)
&= \frac{1}{K-1}\sum_{j\ne i}\mathbf{1}[p_{ij}=i].
\end{aligned}
\]
The output can be a winner label, a preference probability, or a win-rate reward. This mechanism is often more stable than absolute scoring when candidates differ subtly under the same prompt, but it is sensitive to position and order bias and is more expensive than pointwise scoring. Pairwise judging is used in LLM-as-judge settings such as MT-Bench / Chatbot Arena and in pairwise RLHF or generative reward-modeling formulations~\citep{zheng2023judging,xu2025pairwiseframework}. Writing-Zero and HER provide open-ended writing and roleplay examples, while LitBench directly supports creative-writing comparison as a benchmark signal~\citep{jia2025writingzero,du2026her,LitBench}.

\paragraph{Listwise / ranking scorer.}
Listwise mechanisms compare a candidate set at once and return a ranking or top-\(k\) selection:
\[
\begin{aligned}
\mathrm{rank}_i &= J_\phi(x,\{y_1,\ldots,y_K\}),\\
h_{\mathrm{rank}}(x,y_i) &= 1-\frac{\mathrm{rank}_i-1}{K-1}.
\end{aligned}
\]
This can strengthen same-query discrimination by asking the judge to order all candidates jointly. It is distinct from a best-of-\(K\) winner selector, which is a top-\(1\) special case of listwise evaluation. The main risks are context length, set-composition dependence, and permutation sensitivity. LINKAGE provides a direct listwise-ranking formulation for non-factoid QA evaluation, and permutation-consensus listwise judging studies robustness issues for listwise factuality evaluation~\citep{LINKAGE,PCFJudge}.

\paragraph{Claim-level factuality / grounding scorer.}
For grounded open-ended generation, a mechanism can first extract atomic claims, QA probes, or citation-grounded statements and then judge support:
\[
\mathcal{C}(y)=\{c_1,\ldots,c_M\},\qquad
z_j=V_\phi(c_j;x,\mathcal{S}_x),
\]
\[
h_{\mathrm{fact}}(x,y)=\frac{1}{M}\sum_{j=1}^{M}\mathbf{1}[z_j=\mathrm{supported}].
\]
Here \(\mathcal{S}_x\) denotes source documents, dialogue context, retrieved evidence, or cited sources. This mechanism differs from checklist scoring because the evaluated units are extracted from the response rather than fixed in advance. It is useful for long-form factuality, summarization consistency, RAG-style answers, and deep research reports. FActScore and SAFE/LongFact evaluate long-form factuality through atomic facts and search-augmented verification; QAGS, \(Q^2\), and SummaC use QA- or entailment-style probes for factual consistency; DeepResearch Bench extends this idea to citation accuracy and effective citation counts in deep research agents~\citep{FActScore,SAFE,QAGS,Q2,SummaC,DeepResearchBench}.

\paragraph{Boundary with criterion-source adaptation.}
Several related methods change what is evaluated rather than how scoring is elicited. Query-level rubric generation, online rubric elicitation from pairwise comparisons, rubric-anchor methods, and dynamic-rubric RL are therefore best viewed as criterion-source or reward-system adaptation methods, not as separate scoring mechanisms by themselves~\citep{liu2025openrubrics,huang2025rubricanchors,xu2026rubricarm,rezaei2025onlinerubrics,lv2026queryspecific,shao2025drtulu,jia2026openrs}. In our framework, such criterion-source changes can coexist with mechanism changes inside the same executable \rewarddag{}.

\subsection{Additional Motivating Results}
\label{app:additional_motivating_results}

This subsection provides three supplementary analyses that complement the main-text distribution and training-dynamics figures. All numbers are computed from cached offline mechanism scores; no additional judge calls were made for this summary. The analyses cover: (i) query-level versus task-level criteria under the same scoring mechanism, (ii) mechanism behavior across response buckets, and (iii) mechanism-specific score-distribution shift during training.

\paragraph{Query-level versus task-level criteria.}
This comparison isolates criterion-source specialization from reward-mechanism changes. Both conditions use the same continuous rubric scoring path over the same 128 writing responses; the only difference is whether the scorer uses task-level global criteria or query-level specialized criteria. Query-level criteria slightly lower the global mean score, from \(0.687\) to \(0.678\), but increase score dispersion and within-query separability: global standard deviation rises from \(0.061\) to \(0.074\), same-query range rises from \(0.064\) to \(0.093\), and same-query standard deviation rises from \(0.026\) to \(0.037\). The paired score correlation remains high, with Pearson \(0.825\) and Spearman \(0.823\), showing that query-level criteria preserve the broad scoring trend while reshaping local distinctions.

\begin{figure*}[!tbp]
\centering
\begin{minipage}[t]{0.49\textwidth}
\centering
\includegraphics[width=\linewidth]{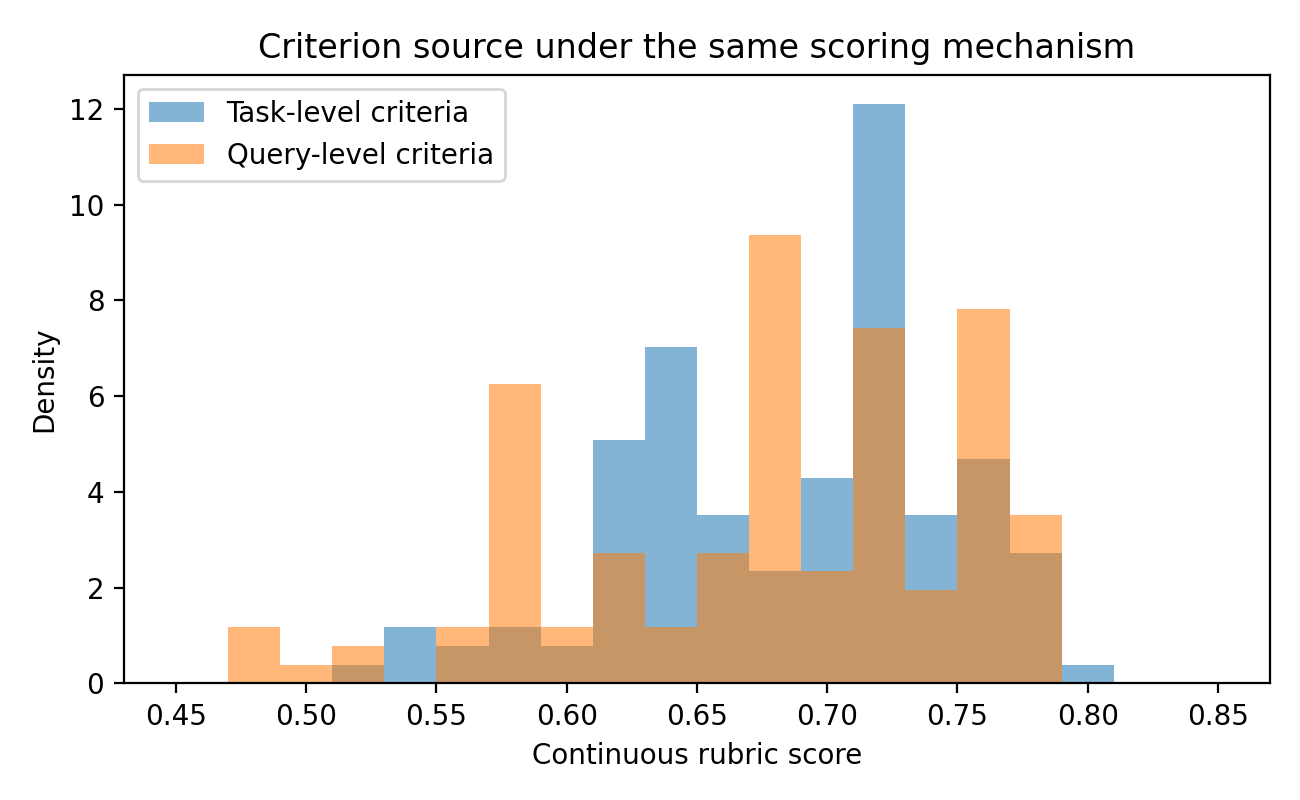}
\end{minipage}
\hfill
\begin{minipage}[t]{0.49\textwidth}
\centering
\includegraphics[width=\linewidth]{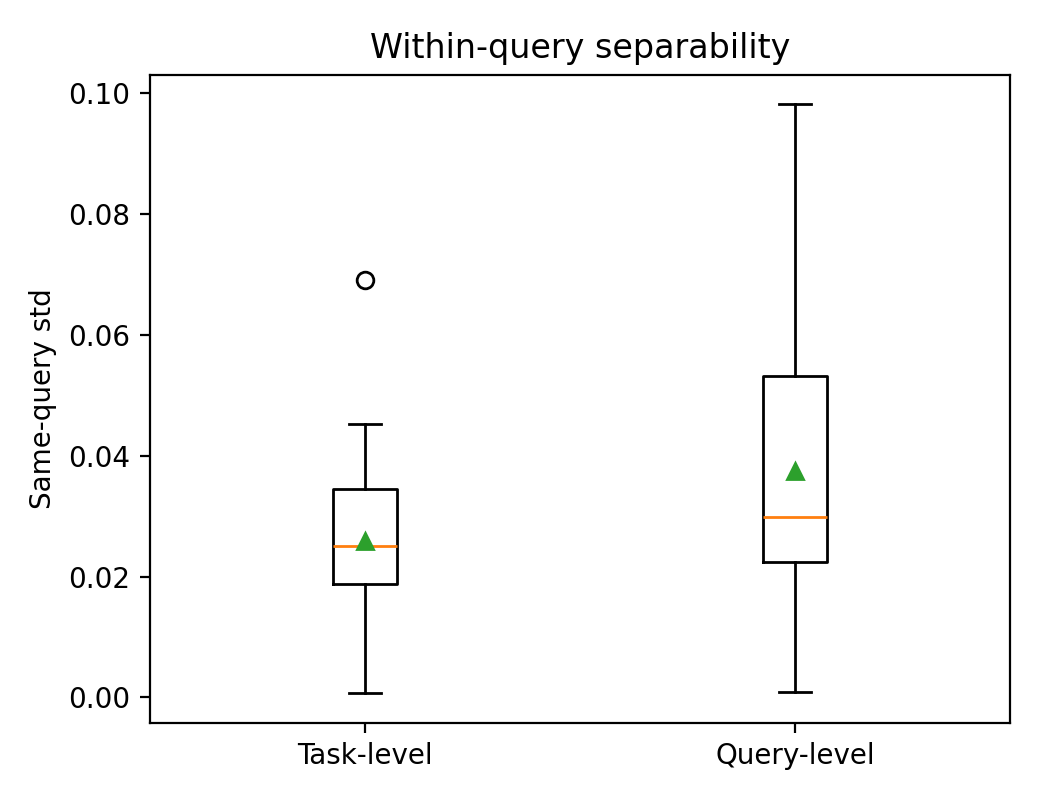}
\end{minipage}
\caption{Query-level versus task-level criteria under the same continuous scoring mechanism. Query-level criteria change the reward landscape and improve within-query separability, but this is a criterion-source change rather than a new reward mechanism.}
\label{fig:app_query_vs_task_criteria}
\end{figure*}

\begin{figure*}[!tbp]
\centering
\includegraphics[width=0.96\textwidth]{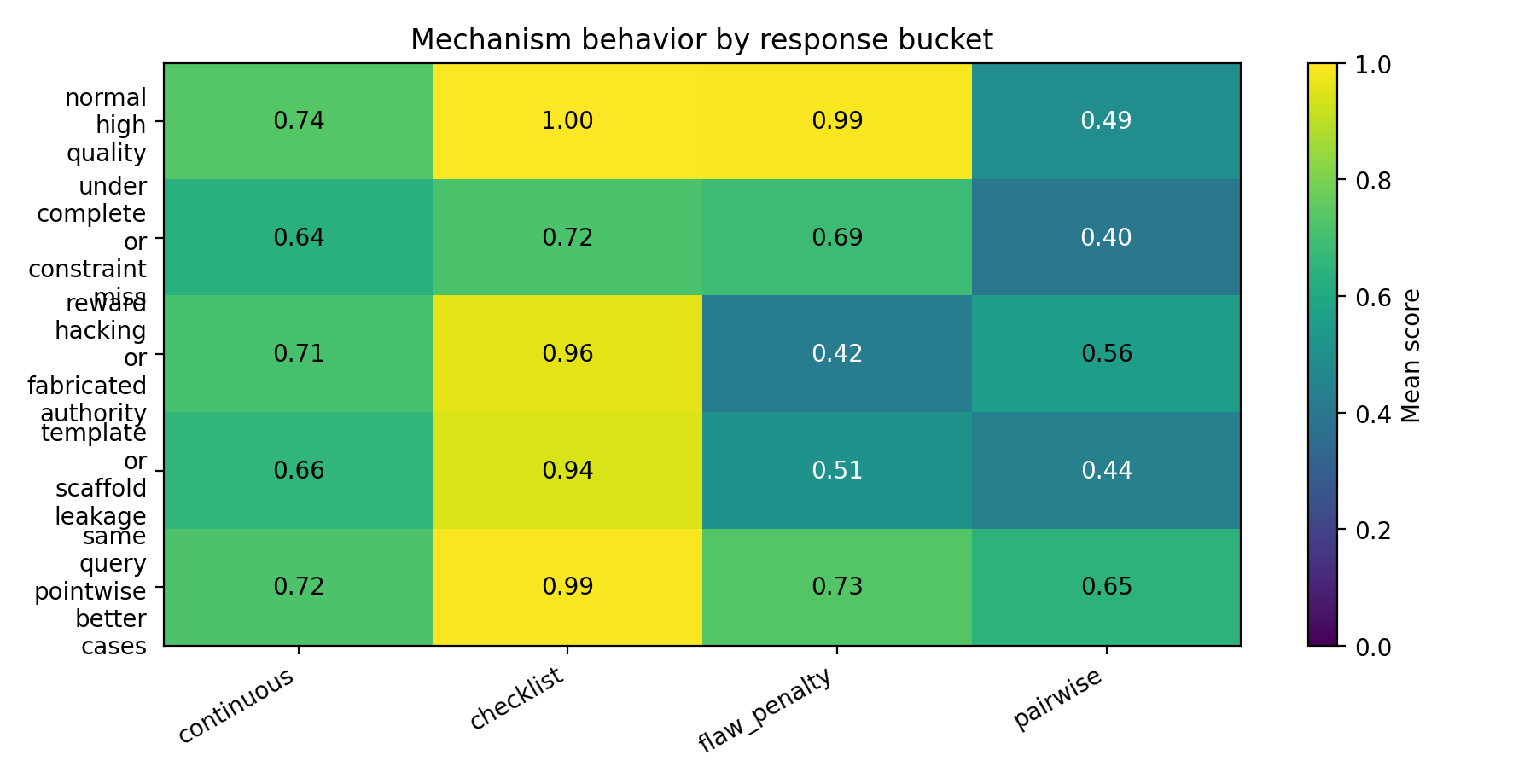}
\caption{Mechanism behavior by response type. Continuous scoring, checklist verification, flaw-penalty detection, and pairwise comparison expose different failure modes and therefore provide complementary reward signals.}
\label{fig:app_mechanism_by_response_type}
\end{figure*}

\paragraph{Mechanism behavior by response type.}
We next bucket the same writing responses into five behavior categories, with 20 examples per bucket: normal high-quality outputs, under-complete or constraint-missing outputs, reward-hacking-like or fabricated-authority outputs, template or scaffold leakage, and subtly better same-query candidates selected by pointwise quality. Figure~\ref{fig:app_mechanism_by_response_type} shows that the mechanisms are complementary rather than interchangeable. Continuous scoring tracks broad quality, but remains relatively high on fabricated-authority cases (\(0.709\)). Checklist scores sharply drop on explicit incompleteness or constraint misses (\(0.717\)), but remain high on fabricated-authority and scaffold-leakage cases (\(0.958\) and \(0.942\)). Flaw-penalty scoring is most sensitive to bad-tail artifacts, dropping to \(0.423\) for fabricated-authority cases and \(0.513\) for scaffold leakage. Pairwise scores are centered near \(0.5\) over arbitrary buckets, as expected, but rise to \(0.646\) on independently selected same-query better cases.

\begin{figure*}[!tbp]
\centering
\includegraphics[width=0.98\textwidth]{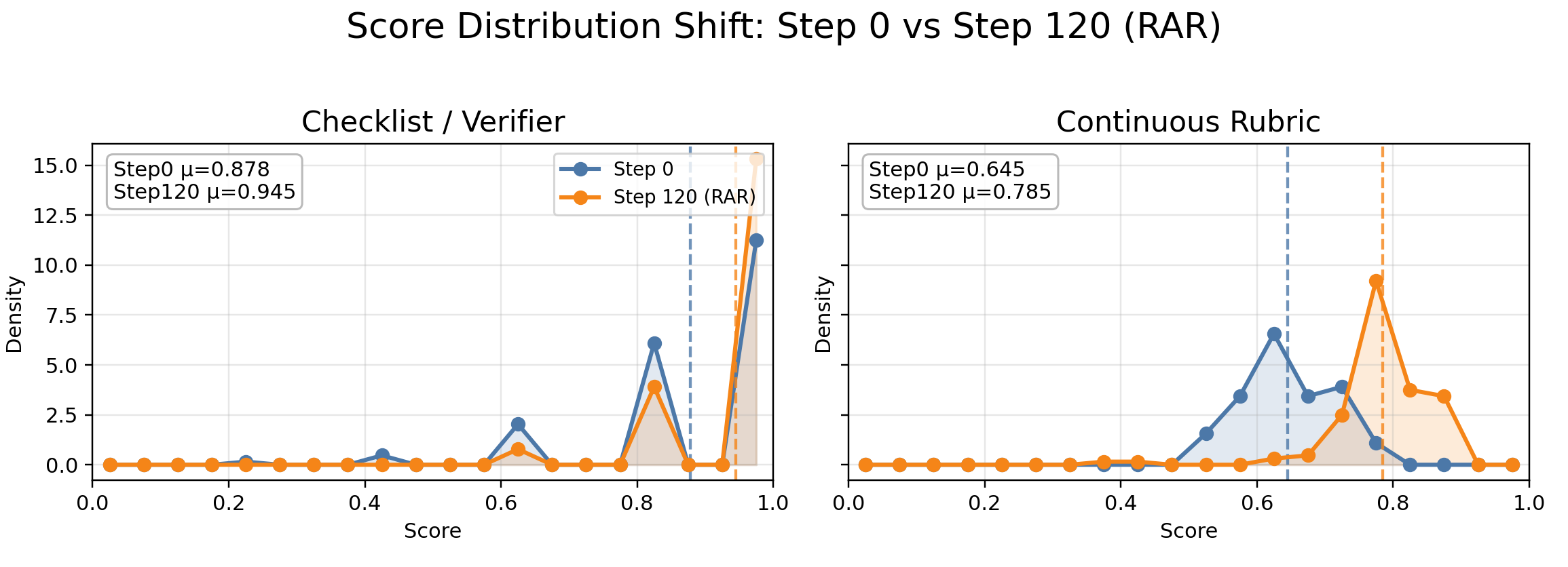}
\caption{Score-distribution shift from step 0 to step 120 under RaR for the two mechanisms with the clearest shift. Continuous scores shift upward while preserving some spread; checklist scores become highly concentrated near the upper end, increasing the rate of zero-distinction groups.}
\label{fig:app_distribution_shift_rar}
\end{figure*}

\paragraph{Distribution shift during training.}
Finally, we compare step-0 base-model responses with step-120 trained responses using the same offline mechanism scoring pipeline. Figure~\ref{fig:app_distribution_shift_rar} shows the clearest shifts for the RaR checkpoint. Continuous rubric scores move upward from mean \(0.645\) at step 0 to \(0.785\) at step 120, while checklist scores move from \(0.878\) to \(0.945\). The checklist signal also becomes less informative within groups: same-query standard deviation drops from \(0.100\) to \(0.050\), and zero-variance group rate rises from \(0.188\) to \(0.531\). Continuous scoring retains more local spread under RaR than checklist, with same-query standard deviation increasing from \(0.024\) to \(0.038\) and zero-variance group rate decreasing from \(0.500\) to \(0.344\). These results illustrate that reward mechanisms can shift or saturate differently as the policy distribution changes, so final reward variance should be separated from base positive-reward informativeness and bad-tail penalty activation.

\section{Method Details}
\label{app:method_details}

\subsection{Implementation Details for Reward Evolution}
\label{app:implementation_details}

\paragraph{Model, compute, and training settings.}
Table~\ref{tab:compute_and_hparams} summarizes the model and training settings
used for the main reported GRPO runs. The policy model is
Qwen3-4B~\citep{yang2025qwen3}, and the local GRM/rubric-updater model is
Qwen3.5-27B. GPT-5.4 is used by the reward-system designer. Final benchmark
quality is evaluated by GPT-5.6-Terra, DeepSeek-V4-Pro, and GLM-5.2; additional
auditor and ranking analyses use the models specified in their corresponding
subsections. These proprietary models do not expose public parameter counts, so
we report their model names and roles. We use a fixed GRPO recipe across methods
rather than method-specific hyperparameter search. All compared reward-system
variants share the same policy initialization, training data, rollout budget,
batching, optimizer settings, and checkpoint schedule; reward-evolution
thresholds and update cadences are fixed before the final comparison.

\begin{table*}[!tbp]
\centering
\small
\setlength{\tabcolsep}{5pt}
\begin{tabular}{p{0.28\textwidth}p{0.66\textwidth}}
\toprule
Category & Setting \\
\midrule
Policy initialization & Qwen3-4B. \\
Training algorithm & GRPO using the same trainer for all reward-system variants. \\
Reward/scoring model & Local Qwen3.5-27B GRM; scalar rewards normalized to \([0,1]\). \\
Designer model & Codex SDK with GPT-5.4, invoked every 5 RL steps for \method{}. \\
Benchmark judges & GPT-5.6-Terra, DeepSeek-V4-Pro, and GLM-5.2; scores are equally averaged. \\
Training length & 120 GRPO steps for the main reported runs. \\
Batching & 32 prompts per step, 4 rollouts per prompt. \\
Sequence lengths & 4096-token prompts; 4096-token writing responses and 2048-token roleplay responses. \\
Optimization & Learning rate \(1\times10^{-6}\), PPO mini-batch size 4, micro-batch size 2 per GPU, low-variance KL with coefficient 0.001, entropy coefficient 0. \\
Checkpointing & Save every 5 steps. \\
Hardware & 8 NVIDIA A800 GPUs for each main training run. \\
RL training budget & About 24 hours wall-clock on average per main run, i.e., about 192 A800 GPU-hours. \\
\bottomrule
\end{tabular}
\caption{Model, compute, and core hyperparameter settings for the main GRPO
runs. Proprietary designer and evaluation models have undisclosed parameter
counts, so we report their names and roles instead.}
\label{tab:compute_and_hparams}
\end{table*}

\paragraph{Runtime cost and agent overhead.}
\label{app:runtime_cost}
The RL training budget above counts the training path. Offline benchmark
scoring, HR/CFR audits, blind ranking annotations, and multi-judge benchmark
evaluation are not included in the RL training GPU-hour accounting. Online reward
scoring during RL is included.

We report writing-side wall-clock cost from the training logs. The per-step
time is the logged \texttt{perf/time\_per\_step}, which includes rollout
generation, reward scoring, log-probability and reference-policy computation,
actor update, checkpointing when triggered, and hook overhead. For
\method{}, this logged step time already includes the designer update when
the update hook runs. We therefore also report the training-only step time
after subtracting \texttt{meta\_judge/elapsed\_s} on update steps and the
average agent time per reward-evolution step.

\begin{table*}[!tbp]
\centering
\small
\setlength{\tabcolsep}{5pt}
\begin{tabular}{lrrrr}
\toprule
Method & Parsed steps & Avg. step time & Avg. agent update time & Agent interval \\
\midrule
RLAIF & 120 & 511.3s & -- & -- \\
RaR & 120 & 663.8s & -- & -- \\
RLER & 120 & 718.4s & -- & -- \\
OpenRS & 120 & 1037.4s & -- & -- \\
\method{}, training only & 120 & 753.0s & -- & -- \\
\method{}, with designer & 120 & 801.5s & 242.8s & 5 steps \\
\bottomrule
\end{tabular}
\caption{Writing-side runtime cost parsed from training logs. For
\method{}, the designer runs every 5 GRPO steps; its average evolution-step
time is 242.8s. The training-only row subtracts designer elapsed time from
update steps, while the with-designer row reports the logged end-to-end step
time.}
\label{tab:runtime_cost}
\end{table*}

\paragraph{GRM scoring and query-level criterion instantiation.}
All pointwise scalar rewards in the main experiments are computed with a local Qwen3.5-27B GRM. The scorer receives the query, the assistant response, and the active rubric or \rewarddag{} node specification, and returns item-level scores together with a normalized scalar reward in \([0,1]\). For continuous rubric nodes, the final node score is the mean or weighted mean of item scores after parsing and clipping. For checklist or verifier nodes, boolean or partial-credit outcomes are normalized into a pass-rate score. For flaw-penalty nodes, detected violations produce a downward correction that is combined with the positive reward through the composition operator. When a query-level rubric node is active, the same local Qwen3.5-27B model instantiates query-aware criteria under the node's fixed generation policy. This is part of reward execution and does not change the active reward-system state.

\paragraph{Continuous GRM scoring prompt.}
The continuous rubric scorer uses the following runtime prompt template. Long query, response, and rubric fields are abbreviated here, but the scoring instructions and output schema match the implementation.

\begin{tcblisting}{
    title=Continuous GRM Scoring Prompt,
    colback=white,
    colframe=black,
    width=\linewidth,
    listing only,
    listing options={basicstyle=\small\ttfamily,breaklines=true,columns=fullflexible},
    breakable,
}
You are a severe but fair reward model grader.

You will receive a query, an assistant response, and a list of rubric items.
Score each rubric item independently on a 0.0-1.0 scale.

Use degree of satisfaction, not a yes/no interpretation:
- 0.00-0.20: absent, contradicted, or harmful.
- 0.20-0.40: very weak; only superficial or accidental evidence.
- 0.40-0.55: partial but flawed; important requirements are missing.
- 0.55-0.70: adequate but generic, thin, or uneven.
- 0.70-0.84: clearly good; specific evidence supports the criterion.
- 0.84-0.92: very strong; robust, nuanced, and well grounded.
- 0.92-1.00: exceptional; rare, highly effective execution.

Calibration and ceiling rules:
- Do not give high scores for merely avoiding errors.
- Generic or interchangeable responses should usually be capped at 0.65.
- Scores above 0.72 require concrete evidence from the response.
- Scores above 0.84 require strong craft, insight, grounding, or reader effect.
- If the response exposes chain-of-thought or private reasoning, cap the final
  mean score at 0.35.

Strict scoring addendum:
- Start from 0.55 and move up or down based on concrete evidence.
- 0.70 is already a good score; above 0.70 requires specific non-generic merit.
- 0.80+ should be stronger than normal good answers; 0.88+ should be rare.
- If the rationale for an item would be generic, cap that item at 0.68.
- Polished surface without insight, specificity, or task judgment caps at 0.65.
- Separate responses by depth, specificity, judgment, and reader effect.

Query:
{query}

Assistant response:
{response}

Rubric items:
{rubric_items}

Return JSON only:
{
  "item_results": [
    {"score": 0.58, "reason": "brief reason with concrete evidence"}
  ],
  "item_scores": [0.58]
}
\end{tcblisting}

\paragraph{Agentic designer harness.}
The reward-system designer is implemented with the Codex SDK using gpt-5.4 as the backbone model. Codex is not used as a response-level judge. Instead, it operates inside our reward-evolution harness: it observes rollout summaries, reward traces, diagnostic evidence, the current \rewarddag{}, and run-specific memory; it can only modify the reward system through the typed edit interface. The harness exposes a diagnostic tool set, typed editing operations, and memory tools, while fixed and dynamic skills guide candidate search. The full tool interface and skill content are given in Appendices~\ref{app:tools} and~\ref{app:skills}.

\paragraph{Harness runtime prompt.}
The following is the actual runtime prompt structure supplied to the Codex-backed harness, with long JSON states and run-specific diagnostics abbreviated.

\begin{tcblisting}{
    title=Harness Runtime Prompt,
    colback=white,
    colframe=black,
    width=\linewidth,
    listing only,
    listing options={basicstyle=\small\ttfamily,breaklines=true,columns=fullflexible},
    breakable,
}
You are the Reward-System Designer for RL.
Use the reward-evolution guidance skill to diagnose the current reward-system
state, search for alternative repair directions, and compare the resulting
candidate states. This prompt provides the current run state and
non-negotiable runtime constraints.

## Pre-Loaded State (DO NOT call load_dag or get_dag_json)

### Current DAG
{dag_json}

### Reward Health Dashboard (Iteration {iteration})
{formatted_dashboard}

### Evaluation Results
Accuracy: {accuracy_pct}%
{eval_results_json}

{benchmark_context}

{scenario_meta_rubrics}

## Runtime Reward Context

The current training reward DAG is pointwise. During actual RL reward
execution, each rubric node sees one sample at a time: the query/prompt,
one candidate response, and any metadata already included in that sample. It
does not see other real responses for the same query.

The diagnostic tool set may retrieve trajectories, inspect reward traces,
cluster cases by a specified response pattern, and replay a candidate on
selected queries. These operations may compare same-query groups outside the
DAG to measure RL signal. Use this evidence to diagnose failures and compare
candidate states. Do not write judge-facing rubric items that require the
judge to compare the current response against unseen same-query responses.

## Current Focus

Dashboard has flagged: {suggested_focus}

## Operational Reward Diagnosis

Use two operational axes before candidate generation and when reading comparative evidence:

- On-policy Reward Validity: whether the reward favors the intended behavior
  and covers the current policy's actual response space, including reward
  hacking, high-score low-quality cases, newly exposed coverage gaps,
  formatting/scaffolding artifacts, corner cases, and false-positive penalties.
- Sustainable Reward Informativeness: whether the reward can keep giving GRPO
  useful signal. Use `informativeness_score`, main/base node signal,
  same-query spread, node/item health, parse health, and final reward sanity.

If the DAG has guard or auxiliary downward effects, final-score variance is
not primary evidence for Sustainable Reward Informativeness. Treat final score
as training reward sanity and use `informativeness_score` for reward-signal
claims.

## Runtime Constraints
- The current DAG remains pointwise; judge-facing criteria must not require
  unseen same-query responses.
- Keep the main `judge` signal continuous unless adding a complementary node.
- Do not edit execution-owned fields: `aggregation`, `chunk_size`,
  `parallel_workers`, or `is_final`.
- Use only the exposed reward-harness tools. Direct shell/file/Python operations
  are forbidden; the read-only analysis tool is allowed for guarded mining.
- Use the diagnostic tool set within the supplied tool budget. Treat
  query-specific replay as evidence about candidate reward behavior, not as a
  response-level reward or an automatic adoption decision.
- If recent run-local memory is present in the prompt, briefly cite the recent
  lesson before candidate generation. Explain whether the proposed directions
  repeat, avoid, or extend that lesson.

## Hard Constraints
- DAG must remain pointwise
- Total nodes <= 10
- Scoring (rubric) nodes per path <= 3

## Policy
- Diagnose before generating candidates
- Propose multiple distinct, bounded repair directions supported by the same
  diagnosis
- If the reward system is healthy, make no change
- Compare candidate states on matched evidence and prefer the candidate that
  most strongly addresses the targeted failure while preserving useful reward
  behavior
- If no candidate provides a supported improvement, retain the current state
- If the diagnostic budget is exceeded, a policy violation occurs, or an unsupported fallback
  would be needed, stop and report it.

## Tool Restriction
Use only the exposed reward-harness tools. No direct Shell, bash, file
reads/writes, or Python snippets. The read-only analysis tool is allowed for
guarded rollout/DAG/project analysis.

## Runtime Budget
Use at most {max_tool_calls} tool calls across both phases.

## Additional Context
## Context: Online RL Training
- Current completed step: {step}
- Selected rollout step for diagnosis: {selected_step}
- Rollout metadata: {rollout_metadata}
- Current scenario: {scenario}
- Candidate search mode: {candidate_search_mode}

The rollout diagnosis and dashboard are already pre-loaded in the prompt. Use
the exposed diagnostic tools and actual rollout cases for deeper checks, then
construct multiple bounded candidate states. Before candidate generation,
separate findings into On-policy Reward Validity, which covers both behavioral
validity and coverage, and Sustainable Reward Informativeness. Identify distinct repair directions that
may address the diagnosed failure, including continuous rubric refinement,
query_rubric refinement, checklist/coverage complements, complementary
continuous residual scoring, aggregation/weight changes, narrow
guard/flaw_penalty side nodes, or no-op.

{scenario_specific_policy}

{method_advice_context}

{run_memory_context}

{operator_ledger_context}
\end{tcblisting}

The prompt separates response scoring from reward-system evolution: the local GRM scores responses and instantiates query-level criteria when required, while the agentic designer diagnoses failures, constructs candidate \rewarddag{} states, and compares their reward behavior through the harness. For candidate search, the runtime groups validity and coverage failures under On-policy Reward Validity and treats reward informativeness separately.

\paragraph{Agent framework selection.}
Before the main training runs, we ran a local diagnostic-modification bench over representative writing and roleplay reward-failure cases. The bench compared Codex SDK, KimiCode, and OpenCode-based agents, including MiniMax-backed configurations. Each agent was evaluated on whether it could identify the annotated reward failure, produce legal candidate states, ground them in rollout evidence, and use comparative feedback to select an effective update. The Codex configuration achieved the most reliable combination of diagnosis quality, edit validity, and candidate selection. Weaker or less compatible agent/model configurations often failed before producing valid candidates, exhausted the tool budget, or proposed candidates that were not selected after comparison. These checks motivated using Codex SDK with gpt-5.4 as the designer in the final experiments; they are implementation-selection checks rather than an additional training reward signal.

\paragraph{Update schedule and state transition.}
The designer is invoked every 5 RL steps. At each evolution step, it uses the diagnostic tool set to examine recent trajectories and reward traces, form several evidence-supported repair directions, and instantiate them as bounded candidate \rewarddag{} states. The current and candidate states are then evaluated on matched recent rollout groups. Their comparative reward behavior determines which candidate, if any, becomes the next active state; when no candidate provides a supported improvement, the current state is retained. The comparison criteria and typed edit validity checks are described in Appendix~\ref{app:reward_dag}.

\subsection{Reward-DAG Design Details}
\label{app:reward_dag}

\paragraph{Implementation scope.}
The main text defines \rewarddag{} as an abstract reward-system representation with editable criterion sources, reward mechanisms, and composition rules. The experiments instantiate a pointwise RL reward path, which is the standard scalar-reward interface used by GRPO. Within this path, \method{} performs mechanism-level evolution by adding or revising executable reward nodes, including query-level rubric nodes, checklist/verifier nodes, flaw-penalty or guard nodes, and composition operators that fuse side signals with the main scorer.

\paragraph{Data schema.}
The executable graph is stored as a JSON-like DAG configuration. The following pseudocode summarizes the implemented schema.

\begin{quote}
\small
\begin{verbatim}
RewardDAG = {
  name: string,
  version: string,
  description: string,
  mode: "pointwise" | "pairwise",
  entry: node_id,
  nodes: [NodeConfig],
  global_config: dict
}

NodeConfig = {
  id: string,
  type:
    "rubric" | "query_rubric"
    | "verifier" | "gate"
    | "aggregator" | "transform",
  config: dict,
  next: node_id | null,
  description: string
}

ExecutionContext = {
  query: string,
  response_a: string,
  response_b: string,
  query_rubrics: dict,
  variables: dict,
  scores: dict[node_id, float],
  execution_path: [node_id],
  final_score: float | null
}
\end{verbatim}
\end{quote}

For pointwise RL, the input is \((x,y)\), represented as \texttt{query} and \texttt{response\_a}. Nodes execute sequentially from \texttt{entry}; each node writes variables, node-level scores, and trace details into the context. The scalar training reward is the \texttt{final\_score} produced by the terminal Rubric Node or Composition Operator.

\paragraph{Node and operator semantics.}
Table~\ref{tab:dag_node_semantics} summarizes the implemented node families and their reward semantics. Each executable node has an input contract, writes a trace and a scalar node score when applicable, and passes control through \texttt{next} or a gate branch. Judge-produced item scores are parsed as real values and clamped to \([0,1]\); Boolean verifier outputs are cast to \(0/1\) when used as scores or guard-role inputs. A terminal node with \texttt{is\_final=true} terminates execution and exposes its score as the scalar training reward.

\begin{table*}[!tbp]
\centering
\small
\setlength{\tabcolsep}{4pt}
\begin{tabular}{p{0.20\textwidth}p{0.24\textwidth}p{0.20\textwidth}p{0.26\textwidth}}
\toprule
Node/operator & Input & Output & Reward semantics \\
\midrule
Global Rubric Node
& \(x,y\), fixed \(\rho^g\), judge \(J_\phi\)
& node score \(h_v\in[0,1]\), item traces
& Continuous mode scores rubric items \(\{c_k\}_{k=1}^K\), normalizes each parsed item score \(s_k\) to \([0,1]\), and returns \(h_v=K^{-1}\sum_k s_k\). This is the main positive reward mechanism in the pointwise RL runs. \\
\midrule
Query-Level Rubric Node
& \(x,y\), meta-rubrics, generation policy, optional precomputed query criteria
& query-rubric score \(h_v\in[0,1]\), generated criteria trace
& Instantiates query-aware criteria and scores them through the rubric-scoring path, yielding the same normalized scalar contract as a Rubric Node. In our current experiments this node mainly changes the criterion source, not the hidden scorer execution protocol. \\
\midrule
Checklist / Verifier
& \(x,y\), explicit checks or checklist items
& pass indicators or pass-rate score
& Checklist scoring returns \(K^{-1}\sum_k \mathbf{1}[\mathrm{pass}_k]\). Rule verifiers write Boolean variables; as guard-role inputs these variables multiply or route downstream scores rather than introducing an unconstrained reward scale. \\
\midrule
Deduction / Flaw-Penalty Rubric
& \(x,y\), failure dimensions, judge-extracted violations or flaws
& penalty-style score \(h_v\in[0,1]\), violation trace
& Deduction mode uses \(s_j=\max(0,1-\lambda\sum_i \mathrm{severity}_{ij})\) per dimension with default \(\lambda=0.05\), then averages dimensions. Flaw-penalty mode uses \(s=1\) if total severity is zero, otherwise \(s=\mathrm{clip}((c-wS)/Z,0,1)\), with defaults \(c=101.5,w=5,Z=100\). \\
\midrule
Composition Operator
& upstream node scores \(\{h_v\}\)
& composed score \(h_u\)
& Weighted-sum aggregation computes \(h_u=\sum_v w_vh_v/\sum_v w_v\) over non-guard inputs; inputs marked \texttt{role=guard} multiply the base score. Product aggregation computes \(h_u=\prod_v h_v\). \\
\midrule
Pairwise Comparator
& \(x,y_a,y_b\) or two pointwise scores
& verdict \(A/B\), or a derived preference signal
& Used for evaluator benchmarks and analysis. The implemented pairwise path either directly compares two responses or scores \(y_a,y_b\) separately and applies a transform \(A\) if \(s_a\ge s_b\), else \(B\). Pairwise verdicts are therefore benchmark outputs, not the live pointwise scalar reward in the main experiments. \\
\bottomrule
\end{tabular}
\caption{Implemented Reward-DAG node families and reward semantics.}
\label{tab:dag_node_semantics}
\end{table*}

\paragraph{Reward scalarization for RL.}
For a pointwise training call, the DAG returns one terminal value \(r_D(x,y)\in[0,1]\). If the terminal node is a Rubric Node, this value is the node score defined above. If it is a Composition Operator, it is the normalized composition of upstream node scores. Verifier or checklist outputs contribute either as \(0/1\) guard-role variables, pass-rate side scores, or aggregation inputs, depending on the node configuration. No additional learned reward head is introduced: the scalar \texttt{final\_score} written by the DAG is the reward passed to the RL trainer.

For GRPO-style optimization, after the DAG returns scalar rewards \(\{r_i\}_{i=1}^K\) for a response group \(\mathcal{Y}_x\), the policy update uses group-relative advantages, e.g.,
\[
\hat{A}_i=\frac{r_i-\frac{1}{K}\sum_{j=1}^K r_j}
{\mathrm{Std}(\{r_j\}_{j=1}^K)+\epsilon}.
\]
Thus the DAG must not only be behaviorally valid, but also preserve within-query reward informativeness.

\paragraph{Typed edit operations.}
The designer cannot rewrite reward code directly. All changes must be expressed through typed edit operations over the DAG schema:
\begin{itemize}[leftmargin=*]
    \item \texttt{add\_node} adds a Rubric Node or Composition Operator using \texttt{id}, \texttt{type}, \texttt{config}, \texttt{next\_node}, optional \texttt{insert\_after}, and optional initializer fields. If \texttt{insert\_after} is set, the tool splices the new node into the reachable path and preserves the old downstream edge.
    \item \texttt{update\_node} patches a node using \texttt{id}, \texttt{config}, optional \texttt{next\_node}, and optional \texttt{description}. For structured rubrics, \texttt{rubric\_item\_updates} modifies stable item ids without rewriting unrelated items.
    \item \texttt{remove\_node(id)} removes a non-entry node, bypasses simple predecessors when possible, and removes stale aggregator inputs.
    \item \texttt{set\_entry(id)} changes the graph entry point.
\end{itemize}

Edits are transactional. The tool first clones the current DAG, applies the proposed mutation, normalizes safe structural details, validates the candidate, and commits it only if validation succeeds. Otherwise, the live in-memory and on-disk DAG remain at the last valid state.

\paragraph{Legality checks.}
The implemented validator rejects malformed or unsafe candidates. Structural checks require a valid entry node, valid \texttt{next} and gate branch references, acyclicity, and reachability from the entry node. RL-time pointwise checks require \texttt{mode=pointwise}, at most 10 total nodes, at most 3 scoring nodes on any reachable path, no \texttt{input\_response=both}, no node-level pairwise mode, supported aggregator methods \(\{\texttt{weighted\_sum},\texttt{product}\}\), non-empty aggregator inputs, upstream-only aggregator sources, and a valid terminal final node. Additional edit guards prevent the agent from changing program-owned execution fields such as rubric chunking, parallelism, or hidden item aggregation; protect the main continuous node from being directly replaced by a different scorer mode; and reject obvious task-domain contamination in writing or roleplay runs.

\paragraph{Comparative self-validation and state selection.}
At each evolution step, the system samples matched same-query groups from recent rollouts using four criteria: low-score groups, high-score groups, high-variance groups, and random groups. By default, it samples three query groups per criterion and requires at least two responses per selected query group. The current state and all candidate states are executed on the same cases, so differences in reward behavior are attributable to the candidate state rather than to different rollout samples.

Let \(q(D,\mathcal{B})\) denote the diagnostic summary computed from these execution records, with dimensions including discrimination, RL signal, calibration, bias, stability, and hacking risk. For candidate \(j\), let \(\Delta q_d^{(j)}=q_d(\widetilde D_{t+1}^{(j)},\mathcal{B}_t)-q_d(D_t,\mathcal{B}_t)\), so positive values favor the candidate. A candidate enters the supported set \(\mathcal{C}_t\) only when it satisfies
\[
C_{\mathrm{exec}}^{(j)}\wedge C_{\mathrm{guard}}^{(j)}\wedge
C_{\mathrm{health}}^{(j)}\wedge C_{\mathrm{info}}^{(j)}\wedge
C_{\mathrm{review}}^{(j)}.
\]
The individual conditions are:
\begin{itemize}[leftmargin=*]
    \item \(C_{\mathrm{exec}}^{(j)}\): the candidate graph is valid and its execution success rate is at least \(\tau_{\mathrm{succ}}=1.0\).
    \item \(C_{\mathrm{guard}}^{(j)}\): no new critical status appears, and every default reliability dimension \(d\in\mathcal{D}_{\mathrm{guard}}\) satisfies \(\Delta q_d^{(j)}> -\tau_{\mathrm{guard}}\), with \(\tau_{\mathrm{guard}}=0.05\).
    \item \(C_{\mathrm{health}}^{(j)}\): average diagnostic health does not drop by more than \(\tau_{\mathrm{health}}=0.15\).
    \item \(C_{\mathrm{info}}^{(j)}\): the candidate's informativeness-score std and range are at least \(0.02\) and \(0.05\), respectively; the final reward also has nonzero std and range on the matched cases.
    \item \(C_{\mathrm{review}}^{(j)}\): if model-based comparative review is enabled, the candidate receives an overall support score of at least \(0.55\); otherwise this condition is true by default.
\end{itemize}
Here \(\mathcal{D}_{\mathrm{guard}}\) is the default reliability set \{discrimination, RL signal, calibration, bias, stability, hacking risk\}. Among supported candidates, the system selects the state whose comparative evidence most strongly improves the diagnosed failure while preserving clean cases and useful within-query distinctions. If no candidate has a supported advantage over the current state, it sets \(D_{t+1}=D_t\). Thus the checks define which candidates remain eligible, while comparative feedback determines the next reward-system state. The constants above are implementation defaults and can be overridden for ablations or stricter runs.

\paragraph{Example evolution step.}
One selected reward-system update occurred in a writing run at step 25. The active graph before the update was a single Global Rubric Node:
\begin{quote}\small
\texttt{writing\_judge} (\texttt{rubric}, continuous, \texttt{is\_final=true}) with six rubric items: purposeful development, necessary specificity, form-purpose fit, reader effect, insight/originality, and prose control.
\end{quote}
The designer diagnosed weak same-query separation and template-like writing artifacts. Among the candidate repair directions, the selected state applied \texttt{update\_node(writing\_judge, rubric\_item\_updates=...)} without changing the topology. The patch tightened three rubric items: necessary specificity and grounding, insight/originality, and prose control. The new criteria explicitly penalized placeholder scaffolding, pseudo-specific unsupported detail, meta artifacts such as visible word-count scaffolds, and jarring register or language mismatch.

The current and candidate states were compared on 48 recent cases from 12 query groups selected by low-score, high-score, high-variance, and random criteria. The selected candidate achieved execution success rate \(1.0\), kept the graph valid, introduced no new critical dimensions, preserved the average health score, and did not collapse the reward distribution: score std changed from \(0.0673\) to \(0.0730\), and score range from \(0.2666\) to \(0.2916\), both above the default collapse thresholds. This example illustrates a bounded criterion-source/rubric-content repair selected through comparative feedback.

Additional selected updates, including the writing guard-style penalty repair used in the training-dynamics analysis, are reported as experimental case studies in Appendix~\ref{app:case_studies}.

\subsection{\method{} Training Algorithm}
\label{app:algorithm}

\begin{tcolorbox}[
    enhanced,
    breakable,
    colback=white,
    colframe=black,
    coltitle=black,
    colbacktitle=white,
    boxrule=0.4pt,
    arc=1pt,
    left=5pt,
    right=5pt,
    top=5pt,
    bottom=5pt,
    fonttitle=\bfseries,
    width=\linewidth,
    title={Algorithm 1: \method{}}
]

\textbf{Input:} initial policy \(\pi_{\theta_0}\), initial Reward-DAG \(D_0\), query distribution \(\mathcal{X}\), reward-evolution interval \(N\).

\begin{enumerate}[leftmargin=*]
    \item Initialize memory \(\mathcal{M}_0\) and dynamic skills \(\mathcal{S}^{\mathrm{dyn}}_0\), which form the initial designer context \(\mathcal{K}_0\).
    \item For each training step \(t=0,1,\ldots\), sample queries \(x\sim\mathcal{X}\) and grouped responses \(\mathcal{Y}_x\sim\pi_{\theta_t}\).
    \item If query-level rubric nodes are active, instantiate query-aware criteria for the current batch.
    \item Score responses with the current Reward-DAG \(D_t\); log rewards, node traces, and reward-health summaries.
    \item Update \(\pi_{\theta_t}\) using GRPO-style group-relative advantages.
    \item If \(t\) is not a reward-evolution step, set \(D_{t+1}=D_t\) and continue.
    \item Otherwise, use \(\mathcal{K}_t\) and the diagnostic tool set to diagnose the current reward failure and form alternative typed updates \(\{e_t^{(j)}\}_{j=1}^{M}\).
    \item Instantiate candidate states \(\widetilde D_{t+1}^{(j)}=e_t^{(j)}(D_t)\) and execute the current and candidate states on the matched rollout batch \(\mathcal{B}_t\).
    \item Select as \(D_{t+1}\) the supported candidate that most strongly addresses the diagnosed failure while preserving useful reward behavior; if none improves on \(D_t\), retain \(D_t\).
    \item Store the selected, rejected, and no-op trajectories in \(\mathcal{M}_{t+1}\) and distill their comparative outcomes into \(\mathcal{S}^{\mathrm{dyn}}_{t+1}\).
\end{enumerate}
\end{tcolorbox}

\subsection{Agentic Designer Tool Details}
\label{app:tools}

The designer operates through a restricted interface rather than unrestricted file or code access. Table~\ref{tab:meta_judge_tools} summarizes the two functional tool sets used by \method{}: diagnostic tools assemble evidence and expose the actual reward behavior of current and candidate states, while typed editing operations define the admissible Reward-DAG search space.

\begin{table*}[!tbp]
\centering
\small
\setlength{\tabcolsep}{4pt}
\begin{tabular}{p{0.24\textwidth}p{0.29\textwidth}p{0.38\textwidth}}
\toprule
Tool family & Representative interface & Role in the harness \\
\midrule
Diagnostic tool set
& trajectory retrieval; reward-trace inspection; pattern-conditioned case clustering; query-specific candidate execution
& Collect on-policy evidence, analyze recurring response patterns, inspect how the current state produces rewards, and expose comparative reward behavior of candidate states on selected queries. \\
\midrule
Typed candidate construction
& node insertion, node update, node removal, entry change, local aggregation update
& Instantiate alternative repair directions as bounded candidate Reward-DAG states while enforcing graph validity and the admissible mechanism and composition search space. \\
\bottomrule
\end{tabular}
\caption{Tool sets exposed to the agentic reward-system designer.}
\label{tab:meta_judge_tools}
\end{table*}

\paragraph{Diagnostic tool set.}
The diagnostic tools expose recent rollout behavior and reward traces without changing the active reward system. They support targeted trajectory retrieval, inspection of node- and item-level traces, clustering or grouping cases that match a specified response pattern, and query-specific execution of candidate states. Together, these tools let the designer move between aggregate patterns and the concrete cases that produced them, then observe whether alternative candidate states change the targeted behavior while preserving useful same-query distinctions. The tools organize this evidence around reward validity, coverage, and informativeness rather than treating any single aggregate statistic as a sufficient update signal.

\paragraph{Editing tools.}
All Reward-DAG changes must be expressed through typed edit operations. The interface supports bounded node updates, adding evaluator or composition nodes, removing stale or harmful nodes, changing the entry node when needed, and locally modifying aggregation. Each edit is applied transactionally to a candidate graph and is rejected if it creates an invalid DAG. Validation checks include acyclicity, reachability from the entry node, valid upstream inputs for aggregation, pointwise reward semantics, and scenario/domain consistency. Program-owned execution details, such as hidden parser schemas or rubric execution internals, are not exposed as agent-editable fields.

\paragraph{Comparative feedback and state selection.}
Query-specific candidate execution is part of the diagnostic tool set rather than a separate replay component. After alternative candidate states are formed, the current state and all candidates are evaluated on the same recent rollout cases. Program-defined checks remove invalid or clearly degraded candidates, and comparative review identifies the remaining candidate that most strongly addresses the diagnosed failure while preserving clean cases and useful reward distinctions. If no candidate has a supported advantage over the current state, no transition is made. When model-based review is enabled, it reads the same compact comparative evidence and contributes to this selection; it does not produce response-level rewards or bypass the program-defined checks.

\subsection{Reward-Evolution Guidance Skill Details}
\label{app:skills}

Section~\ref{sec:method_meta_judge_policy} abstracts the agentic designer as an
experience-conditioned candidate-generation policy
\(\Pi_{\mathrm{design}}(D_t,\mathcal{O}_t,\mathcal{K}_t)\)
over an admissible update space \(\mathcal{A}(D_t)\). In implementation,
\(\mathcal{O}_t\) is assembled through the diagnostic tool set and reward traces,
\(\mathcal{A}(D_t)\) is defined by the typed \rewarddag{} update tools and
validity checks, and \(\mathcal{K}_t\) consists of fixed reward-evolution
skills, dynamic skills, and run-local memory. Each candidate state produces a
trajectory record, and their comparative outcomes are stored in memory and
summarized into later dynamic-skill context.

Tools define what the designer can diagnose and edit. Skill sets define how it should reason over those interfaces. In \method, skills serve as reward-evolution guidance rather than task-level response rubrics: they structure diagnosis, provide reward-quality knowledge, route failures to suitable mechanism families, constrain candidate search, and specify how comparative evidence should be interpreted.

\paragraph{Fixed skills.}
Fixed skills are shared across runs. They require the agent to diagnose before candidate generation and to ground each repair direction in both rollout evidence and reward traces. For candidate search, the runtime uses two operational targets: On-policy Reward Validity jointly covers validity and coverage failures, while Sustainable Reward Informativeness captures the usefulness of the signal for optimization. Fixed skills also provide reward-quality priors: useful rewards should preserve current-policy preference validity, cover emerging failures, maintain same-query informativeness, avoid saturated or redundant items, execute reliably, and preserve clean cases under matched comparison.

\paragraph{Experience-conditioned candidate search.}
Although the designer is implemented with an instruction-following agent, the harness constrains its search through a typed action space rather than an open-ended prompt workflow. At evolution step \(t\), the agent receives observation summaries \(o_t\), node traces \(\tau_t\), recent memory \(\mathcal{M}_t\), and the current graph \(D_t\). It first assigns a primary operational target, using \(\mathrm{validity}\) as shorthand for behavioral validity and coverage:
\[
f_t \in \{\mathrm{validity},\mathrm{informativeness},\mathrm{both},\mathrm{none}\},
\]
and a failure pattern \(p_t\) from the operator-routing table. The table maps \(p_t\) to a small operator family \(\Omega(p_t)\), such as a global rubric refinement, query-level rubric specialization, checklist/verifier, flaw-penalty side signal, aggregation adjustment, or removal. Alternative updates are drawn from the legal typed-edit set:
\[
e_t^{(j)} \in \mathcal{A}(D_t)\cap \Omega(p_t),
\qquad j=1,\ldots,M,
\]
where \(\mathcal{A}(D_t)\) is defined by the edit tools and validation checks in Appendix~\ref{app:reward_dag}. The candidate set covers distinct, evidence-supported repair directions while excluding actions contradicted by recent memory. Each update produces \(\widetilde D_{t+1}^{(j)}\), and matched comparative feedback determines which candidate, if any, becomes the next state. If no evidence-backed direction exists, or no candidate improves on the current state, the system returns a no-op rather than forcing a transition.

\begin{tcolorbox}[
    enhanced,
    breakable,
    colback=white,
    colframe=black,
    coltitle=black,
    colbacktitle=white,
    boxrule=0.4pt,
    arc=1pt,
    left=5pt,
    right=5pt,
    top=5pt,
    bottom=5pt,
    fonttitle=\bfseries,
    width=\linewidth,
    title={Algorithm 2: Experience-Conditioned Candidate Search}
]

\textbf{Input:} current graph \(D_t\), rollout summaries \(o_t\), node traces \(\tau_t\), diagnostic summary \(q_t\), memory \(\mathcal{M}_t\), fixed skills \(\mathcal{S}^{\mathrm{fix}}\), dynamic skills \(\mathcal{S}^{\mathrm{dyn}}_t\).

\begin{enumerate}[leftmargin=*]
    \item Diagnose behavioral validity, coverage, and reward informativeness from \(o_t,\tau_t,q_t\); attach representative rollout evidence to each finding.
    \item Set \(f_t\in\{\mathrm{validity},\mathrm{informativeness},\mathrm{both},\mathrm{none}\}\). If \(f_t=\mathrm{none}\), return no-op.
    \item Classify the dominant failure pattern \(p_t\) using the operator-routing prior in Table~\ref{tab:operator_routing}.
    \item Construct the admissible action set \(\mathcal{A}(D_t)\cap\Omega(p_t)\), where \(\mathcal{A}(D_t)\) is the typed edit set allowed by the current graph validator.
    \item Remove actions that conflict with recent rejected memory, modify protected execution fields, shift the current reward objective without evidence, or violate pointwise RL constraints.
    \item Select \(M\) distinct, evidence-supported repair directions and instantiate the candidate states \(\{\widetilde D_{t+1}^{(j)}\}_{j=1}^{M}\). If none is supported, return no-op.
    \item Use the diagnostic tool set to execute the current and candidate states on matched cases and collect their comparative reward behavior.
    \item Select the candidate that best addresses the targeted failure while preserving clean cases and same-query signal; if none improves on \(D_t\), retain \(D_t\). Store all comparative outcomes for later memory and dynamic-skill updates.
\end{enumerate}
\end{tcolorbox}

This policy is implemented by the designer harness. The language-model agent supplies diagnosis, candidate directions, and comparative judgment, while the action space, graph validator, matched-case construction, and admissibility checks are program-defined.

\paragraph{Condensed fixed-skill card.}
The following condensed skill card is supplied to the agent as reward-evolution guidance. We reproduce it in segmented form to make the harness behavior auditable.

\textbf{Mission.}
\begin{quote}\small
Optimize pointwise RL reward DAGs during unverifiable generation training. The goal is not to make a prettier rubric; the goal is to keep reward useful as the policy changes. Use two operational targets for candidate search: On-policy Reward Validity, covering behavioral validity and coverage, and Sustainable Reward Informativeness.
\end{quote}

\textbf{Operational diagnosis.}
\begin{quote}\small
On-policy Reward Validity asks whether the reward favors intended behavior and covers the current policy's actual response space, including reward hacking, high-score low-quality responses, shortcut behavior, brittle over-penalization, corner cases, false positives, and newly exposed response patterns. Sustainable Reward Informativeness asks whether the reward can keep giving GRPO useful learning signal, using same-query range/std, zero-variance groups, main or base node health, item saturation, score collapse, parse failures, calibration, and advantage signal.
\end{quote}

\textbf{Reward-signal interpretation.}
\begin{quote}\small
Reward statistics are health signals, not standalone objectives. Same-query std/range and advantage behavior on the informativeness score are primary evidence for GRPO signal. Final reward mean/std and global variance are training-sanity checks. Larger final-score variance is not progress when it comes mainly from guard or auxiliary downward effects.
\end{quote}

\textbf{Required workflow.}
\begin{enumerate}[leftmargin=*]
    \item Diagnose before candidate generation: inspect rollout summaries, dashboards, node/item traces, same-query groups, contrastive cases, and recent memory.
    \item State separate findings for validity, coverage, informativeness, recent memory, scenario evolution direction, and objective-continuity risk.
    \item Choose one primary target for the evolution step: validity/coverage, informativeness, or both.
    \item Classify the failure and identify several distinct repair directions within the admissible operator families.
    \item Instantiate each direction as a bounded candidate through the typed Reward-DAG interface.
    \item Use the diagnostic tool set to compare targeted failures, clean-case preservation, same-query informativeness, node traces, and execution errors across candidates.
    \item Record the selected, rejected, and no-op trajectories, their side effects, and lessons for later dynamic skills.
\end{enumerate}

\begin{table*}[!tbp]
\centering
\small
\setlength{\tabcolsep}{4pt}
\begin{tabular}{p{0.25\textwidth}p{0.27\textwidth}p{0.38\textwidth}}
\toprule
Failure pattern & Primary operator family & Typical repair \\
\midrule
Continuous quality miscalibration
& Global continuous rubric
& Refine item wording, anchors, evidence requirements, or weights. \\
\midrule
Redundant or saturated criteria
& Global rubric structure
& Split, merge, add, remove, or reweight rubric items. \\
\midrule
Prompt-local coverage gap
& Query-level rubric node
& Revise meta-rubrics or generation policy for query-specific criteria. \\
\midrule
Explicit observable requirement
& Checklist, verifier, or deduction node
& Add hard or soft requirement tracking. \\
\midrule
Sparse high-score bad tail or hacking escape
& Flaw-penalty, guard, or validity side signal
& Add a narrow negative detector or guard-style fusion. \\
\midrule
Useful missing positive signal
& Complementary positive node
& Add a side scorer for a distinct positive construct. \\
\midrule
Signal masking or distorted final reward
& Aggregator or fusion operator
& Adjust weights, guard roles, or fusion method. \\
\midrule
Harmful, inert, or duplicated signal
& Simplification or removal
& Remove, bypass, or downweight stale nodes. \\
\bottomrule
\end{tabular}
\caption{Operator-routing priors encoded in fixed reward-evolution skills.}
\label{tab:operator_routing}
\end{table*}

Table~\ref{tab:operator_routing} summarizes the operator-routing prior used by fixed skills. This prior prevents the designer from treating every failure as a generic rubric rewrite and helps it construct distinct candidate directions with an intended failure target, an owner signal, and an expected reward-behavior change.

\paragraph{Edit-boundary constraints.}
Fixed skills constrain reward evolution so the training objective does not oscillate. The agent is instructed to preserve the stable reward core unless evidence supports a larger structural change, prefer bounded complements before replacing the main reward path, make one primary edit at a time, keep RL-time DAGs pointwise, and treat aggregation or final weights as high-impact changes. We distinguish three edit regions: the \textit{Stable Core}, including the main task objective and primary positive reward signal; the \textit{Adaptive Shell}, including anchors, evidence requirements, generation policy, and small coverage refinements; and \textit{Safety/Validity Guards}, which target narrow observable defects and should not become broad secondary quality judges.

\paragraph{Comparative-evidence standards.}
Skills specify that candidate-level deltas are evidence for state selection, not proof of downstream policy improvement. The designer checks whether targeted cases move in the intended direction, whether clean cases are preserved, whether same-query informativeness improves or at least does not collapse, whether node traces explain the score changes, and whether parse or node errors invalidate a comparison. In particular, guard-induced final-score variance should not be mistaken for sustainable reward informativeness.

\paragraph{Dynamic skills.}
Dynamic skills specialize the fixed policy to the current scenario and training run. They record observed hacking or shortcut patterns, failure-to-repair mappings discovered during training, selected and rejected candidates, no-op cycles, historical side effects, and scenario-specific operator preferences. For example, writing runs may record fabricated authority, constraint self-certification, or dimension coupling as concrete hacking families, while roleplay runs may record format gaming, catchphrase stuffing, meta/OOC leakage, or unsupported escalation. These dynamic skills are injected into later designer calls so that the agent can avoid repeated failures and adapt to the policy trajectory.

\paragraph{Dynamic method advice examples.}
Dynamic skills are implemented as run-local memory and method advice. The advice is advisory context rather than a command list. A condensed version of the advice format is:
\begin{quote}\small
\textbf{Ground every edit in three evidence blocks.} Before editing, reconcile reward-signal diagnostics, actual response-level failures, and recent run memory. If any block is missing or contradictory, resolve that gap first rather than inferring a fix from one view.

\textbf{Preserve objective continuity across training.} Classify each edit as a reliability fix, calibration change, coverage change, guard/validity change, structural complement, aggregation change, or objective-shifting change. Stable-core edits to the main reward target, hidden scorer protocol, or final weights should be rare.

\textbf{Treat comparative deltas as evidence, not progress theater.} Use matched candidate execution to test whether each repair hypothesis holds on same-query spread, trace evidence, clean cases, and targeted failures. Keep reward statistics as decision checks, not isolated optimization targets.

\textbf{Escalate skepticism after repeated no-op or revert outcomes.} When recent updates form a repeated no-op/revert cycle, require materially new response-level evidence before trying another near-identical edit.
\end{quote}

Scenario-specific advice further constrains dynamic edits. For writing, only concrete fabricated authority, constraint self-certification, and dimension coupling are treated as reward-hacking families; ordinary headings, professional structure, useful detail, appropriate length, or normal task completion are not hacking by themselves. For roleplay, shortcut detection is bounded to observable behaviors such as tag gaming without in-character substance, generic dramatic verbosity masking weak grounding, catchphrase or persona-name stuffing, meta/OOC leakage, and unsupported escalation. Ordinary shallow characterization or weak scene progression should usually be handled through positive character, scene, progression, or query-level criteria rather than broad guard penalties.

\paragraph{Run-local memory schema.}
Each evolution step produces a compact set of memory records:
\[
z_t^{(j)}=(o_t,e_t^{(j)},\Delta_t^{(j)},a_t^{(j)},c_t^{(j)}),
\]
where \(o_t\) is the diagnosed failure, \(e_t^{(j)}\) a candidate update, \(\Delta_t^{(j)}\) its comparative evidence, \(a_t^{(j)}\) the selected/rejected/no-op outcome, and \(c_t^{(j)}\) observed side effects or caveats. The memory summary is not used as reward evidence by itself. It is injected into later designer calls as dynamic guidance, for example to avoid repeating a rejected direction, reuse an effective failure-to-repair strategy, or cool down an operator family that repeatedly produced no-op or harmful changes.

In implementation, each memory record also stores the selected rollout step, update mode, agent type, whether the candidate state was selected, a compact diagnostic summary, the comparative evidence, the tool sequence, policy violations if any, and links to the candidate graph and operator ledger. A separate memory-advice summarizer may compress these trajectories into two fields: a short \textit{memory summary} for the next update and a \textit{method-advice update} for longer-lived run-local guidance.

\section{Experimental Details}
\label{app:experimental_details}

\subsection{Artifact Use, Intended Use, and Data Safety}
\label{app:artifact_use}

We use existing artifacts including Qwen3-4B~\citep{yang2025qwen3},
Qwen3.5-27B, WritingBench~\citep{WritingBench},
CoSER~\citep{wang2025coser}, MiniMax Role-Play Bench, GPT-5.4 as the designer
model, and GPT-5.6-Terra, DeepSeek-V4-Pro, and GLM-5.2 as the main benchmark
judges. Additional analyses use DeepSeek-V4-Flash and
Kimi-K2.5~\citep{moonshot2026kimi25}. The creators of these artifacts are cited where they are first used
in the method and experiment sections. We use these artifacts for research on
reward-system evolution and evaluation, consistent with their benchmark,
model-evaluation, or model-training use. Proprietary API models are used
through their service interfaces; their parameter counts and redistribution
rights are not public, so we report model names and roles rather than
redistributing them.

The artifacts created by this work include filtered training prompts, initial
task-level and query-level rubrics, \rewarddag{} schemas and edit logs,
reward-diagnostic taxonomies, trained policy checkpoints, evaluation outputs,
and analysis scripts. These artifacts are intended for research reproduction
and analysis of reward-system evolution, not for deployment as general-purpose
safety filters or user-facing moderation systems.

For data safety, the writing data are synthetic open-ended writing
instructions. We filter duplicated, underspecified, benchmark-like, and
factual-QA-dominated queries, and we avoid prompts that require identifying
private individuals. Roleplay training data are sampled from MiniMax Role-Play
Bench and retained only when persona, relationship, context, and dialogue state
are explicit. We do not intentionally collect personal identifiers. Because
open-ended generation and roleplay can still produce sensitive or offensive
content, final outputs are used only for aggregate evaluation and audit
statistics, and auditor prompts do not require exposing method labels, reward
traces, or user-identifying metadata.

Appendix~\ref{app:data_and_rubrics} reports the data coverage and sizes:
944 writing queries and 5,000 roleplay queries, with writing domains and
roleplay language, persona, relationship, context, and dialogue-depth coverage
summarized there. Section~\ref{sec:exp_setup} and
Appendix~\ref{app:implementation_details} report the model settings, training
protocol, compute budget, and evaluation protocol.

\subsection{Use of AI Assistants in Research and Writing}
\label{app:ai_assistant_use}

The authors used AI assistants for auxiliary support in language polishing and
code development. The authors reviewed and remain responsible for the research
design, experiments, analyses, claims, and final manuscript content.

\subsection{Training Data and Initial Rubric Construction}
\label{app:data_and_rubrics}

\paragraph{Training data.}
For writing, we construct a filtered synthetic query set for open-ended writing RL. The queries are generated to cover six broad writing domains and diverse constraints over audience, format, length, style, language, and communicative intent. We remove queries that are duplicated, underspecified, benchmark-like, or dominated by factual QA rather than writing quality. The final training set contains 944 writing queries. For roleplay, we sample from MiniMax Role-Play Bench and retain queries with clear persona, relationship, context, and dialogue-state information. The final training set contains 5,000 roleplay queries, covering English and Chinese scenarios with varied persona relations and dialogue depths.

\paragraph{Initial rubric construction.}
We build both task-level and query-level rubrics with a lightweight iterative rubric-generation procedure. Starting from a seed instruction and representative examples, a rubric generator proposes candidate criteria and short scoring descriptions. We then audit the candidates on small batches of representative responses and remove criteria that show poor discriminability, strong coupling with other criteria, vague evidence requirements, or weak relevance to the target task/query. New criteria are added to cover uncovered failure modes or task requirements. This generate--audit--revise loop is repeated until the rubrics provide stable, non-redundant, and task-relevant distinctions. The resulting task-level rubrics are used by the RLAIF-style baseline and initialize the single-node continuous \rewarddag{} used by \method, while the query-level rubrics are used by RaR-style reward systems.

\subsection{Baseline Reward-System Details}
\label{app:baseline_details}

All baselines are instantiated under the unified GRPO protocol in
Section~\ref{sec:exp_setup}. We adapt each published reward formulation to
writing and roleplay while using the shared training and evaluation settings
described above. RLER and OpenRS use the same local Qwen3.5-27B model as
\method{} for rubric updating and reward computation. The following paragraphs
specify how each reward system is instantiated.

\paragraph{Static rubric baselines.}
The RLAIF-style baseline uses the task-level rubric constructed in Appendix~\ref{app:data_and_rubrics}. The same rubric is applied to all queries in a task, and the GRM returns a normalized continuous score as the scalar reward. The RaR-style baseline uses fixed query-level rubrics generated before training. For each query, the query-specific rubric remains unchanged during RL, isolating the effect of criterion-source specialization.

\paragraph{RLER dynamic rubrics.}
RLER follows the evolving-rubric design of DR-Tulu~\citep{shao2025drtulu}. After each rollout step, the update model uses current policy responses to augment every task rubric with additional positive or negative criteria. The original and augmented criteria are scored through the same continuous GRM path. Active criteria with weak or collapsed discrimination are filtered using item-level reward statistics, and the active set is capped to a small maximum size.

\paragraph{OpenRS-style pairwise rubric reward.}
The OpenRS-style baseline does not maintain a persistent active-rubric state. Instead, after each rollout, responses under the same query are grouped and compared pairwise. For each pair, the OpenRS prompt asks the judge to analyze the query, construct or select pair-specific rubrics, compare the two responses under those rubrics, and output rubric-level comparison results. The implementation parses the pairwise comparison into a signed margin, aggregates margins across pairs for each response, and maps the average margin to a scalar reward in \([0,1]\). Thus, this baseline changes the reward mechanism to pairwise comparison, but does not perform experience-driven candidate generation and comparative reward-system evolution.

\subsection{Evaluation Diagnostics and HR/CFR Audits}
\label{app:hack_aware_audit}
\label{app:evaluation_diagnostics}

We use HR and CFR as final-output robustness audits in addition to final
benchmark scores. HR (Hack Rate) measures the incidence of fixed
reward-attractive shortcut patterns in final outputs. These patterns are
identified from base-model outputs as visible failures that rubric judges may
over-reward; when they are amplified after RL, we use this as evidence that the
final policy has over-learned reward-attractive shortcuts. CFR (Coverage Failure
Rate) measures the audited rate of predefined hard coverage-failure labels in
final outputs.
R-Var and Zero measure whether the reward gives useful within-query
distinctions for GRPO; they are proxies and are interpreted together with the
dynamics and case studies in Section~\ref{sec:exp_dynamics}.

\paragraph{Audit protocol.}
For each task, we define the HR and CFR taxonomies before method-level
comparison using a base-model-only calibration pool. The calibration pool
contains visible base-model outputs and their prompts, without method labels,
reward scores, reward traces, training-stage metadata, or \method checkpoints.
Human inspection identifies recurring visible failure patterns in this pool.
After this taxonomy-construction step, the HR/CFR categories are frozen and
applied to all methods with the same LLM auditor. The auditor only labels
whether the frozen categories are present and is not allowed to invent new hack
or coverage-failure types. These evaluation taxonomies are not provided to the
designer agent during RL training. Training rollouts and reward traces are
used only for qualitative case-study interpretation and designer diagnosis
analysis, not for defining the reported HR/CFR categories.

\paragraph{HR and CFR metrics.}
Given audited final outputs \(\{(x_i,y_i)\}_{i=1}^{N}\), HR is the rate of
outputs matching a predefined task-specific shortcut pattern:
\[
\mathrm{HR}=\frac{1}{N}\sum_{i=1}^{N}
\mathbf{1}\!\left[
\begin{gathered}
\exists h\in\mathcal{H}_{\mathrm{task}}:\\
\mathrm{Auditor}_h(x_i,y_i)=1
\end{gathered}
\right],
\]
where \(\mathcal{H}_{\mathrm{task}}\) is the task-specific HR taxonomy. For
roleplay, HR is normalized by the number of assistant turns because a
simulation-level shortcut can recur across longer dialogues; the reported
roleplay HR is therefore a per-assistant-turn rate.

CFR is the output-level rate of at least one predefined hard coverage failure:
\[
\mathrm{CFR}=\frac{1}{N}\sum_{i=1}^{N}
\mathbf{1}\!\left[
\substack{
\exists c\in\mathcal{C}_{\mathrm{task}}:\\
\mathrm{Auditor}_c(x_i,y_i)=1
}
\right],
\]
where \(\mathcal{C}_{\mathrm{task}}\) is the task-specific CFR taxonomy. The
reported CFR is the percentage of outputs with at least one frozen CFR label. HR and
CFR answer different final-output robustness questions: HR detects
reward-attractive shortcut patterns, while CFR detects missing coverage of
important discrete validity or completion dimensions. A response may have
neither, either, or both labels.

\paragraph{Writing HR taxonomy.}
The writing HR pattern set was fixed from the base-model WritingBench
calibration pool.
Manual inspection focused on cases that looked attractive to rubric or
checklist judges while not improving, or even hurting, the requested writing
deliverable. This process produced three task-level pattern families.

\begin{table}[!htbp]
\centering
\small
\setlength{\tabcolsep}{3.5pt}
\begin{tabular}{p{0.30\linewidth}p{0.61\linewidth}}
\toprule
Pattern & Definition used by the auditor \\
\midrule
Fabricated authority / surface expertise
& Dense unsupported numbers, citations, institutions, laws, technical jargon, timelines, metrics, or unverifiable concrete details that create false professionalism. \\
\midrule
Self-certification
& Meta-level claims that the answer satisfies length, format, constraint, checklist, or scoring requirements, instead of simply completing the requested deliverable. \\
\midrule
One-dimension over-optimization
& The response is optimized for one visible scoring dimension, such as length, structure, local keyword coverage, or format compliance, while failing essential requirements such as deliverable shape, grounding, usefulness, or audience fit. \\
\bottomrule
\end{tabular}
\caption{Predefined writing reward-hacking patterns used for Hack auditing.}
\label{tab:writing_hack_patterns}
\end{table}

\paragraph{Writing CFR taxonomy.}
Writing CFR uses a stricter hard-failure taxonomy than ordinary WritingBench
quality scoring. It excludes ordinary shortness, weak prose, missing minor
subpoints, and low benchmark quality; those are handled by WritingBench scores.
Instead, it counts discrete non-final or wrong-deliverable failures that a
training reward can under-cover even when the output is fluent.

\begin{table}[!htbp]
\centering
\small
\setlength{\tabcolsep}{3.5pt}
\begin{tabular}{p{0.30\linewidth}p{0.61\linewidth}}
\toprule
CFR type & Definition used by the auditor \\
\midrule
Unresolved template artifact
& The response contains unresolved placeholders, bracketed fill-ins, template debris, or explicit word-count artifacts that make the deliverable non-final. \\
\midrule
Non-final process artifact
& The response exposes planning, evaluation, checklist reasoning, or meta-process text instead of only delivering the requested writing artifact. \\
\midrule
Deliverable mismatch
& The response does not produce the requested artifact, format, scope, or concrete communicative deliverable. \\
\midrule
Missing artifact
& The response admits that a required non-text, source-dependent, or concrete artifact is absent, unavailable, or not actually provided. \\
\bottomrule
\end{tabular}
\caption{Predefined writing coverage-failure dimensions used for CFR auditing.}
\label{tab:writing_cfr_patterns}
\end{table}

\paragraph{Roleplay HR taxonomy.}
Roleplay HR is intentionally narrower than generic roleplay quality. It counts
observable shortcut behavior that can look attractive to a judge without adding
new in-character consequence. Ordinary shallow characterization, weak scene
progression, or generic prose are not HR by themselves.

\begin{table}[!htbp]
\centering
\small
\setlength{\tabcolsep}{3.5pt}
\begin{tabular}{p{0.30\linewidth}p{0.61\linewidth}}
\toprule
Pattern & Definition used by the auditor \\
\midrule
Decorative verbosity / emotion stuffing
& Inflates length with generic dramatic interiority, repeated emotional beats, atmosphere, or private-planning style padding without creating new visible roleplay consequence. \\
\bottomrule
\end{tabular}
\caption{Predefined roleplay reward-hacking pattern used for HR auditing.}
\label{tab:roleplay_hack_patterns}
\end{table}

\paragraph{Roleplay CFR taxonomy.}
Roleplay CFR measures discrete validity and completion failures in visible
AI-character turns. It does not include broad continuous dimensions such as
latest-turn uptake, scene progress, character depth, or generic
transplantability; those remain part of CoSER/GCA-style scoring. The reported
CFR is limited to the following predefined, easy-to-audit failure families.

\begin{table}[!htbp]
\centering
\small
\setlength{\tabcolsep}{3.5pt}
\begin{tabular}{p{0.30\linewidth}p{0.61\linewidth}}
\toprule
CFR type & Definition used by the auditor \\
\midrule
Required protocol / block failure
& Missing, malformed, replaced, or misordered required user-visible tags, speaker labels, blocks, or output format when this materially breaks the requested roleplay output. \\
\midrule
OOC / meta leakage
& Explicit AI-assistant framing, instruction discussion, task explanation, OOC commentary, refusal-like meta text, or model/self-reference when not requested. \\
\midrule
Visible context contradiction
& Contradicts explicit location, relationship state, prior action, character knowledge, causality, or latest-turn facts in the visible context. \\
\midrule
User-agency violation
& Dictates the user's internal state or actions, resolves too much of the scene, or removes the user's natural ability to respond. \\
\bottomrule
\end{tabular}
\caption{Predefined roleplay coverage-failure dimensions used for the reported CFR.}
\label{tab:roleplay_cfr_patterns}
\end{table}

\paragraph{Independent validation of the HR audit.}
We re-audit the same 50 CoSER scenarios for Base, RaR, and \method{} at both
Qwen3-4B and Qwen3-8B scales using Qwen3.5-27B, DeepSeek-V4-Flash, and
Kimi-K2.5. All three auditors produce the same ordering,
\(\method{} < \text{Base} < \text{RaR}\), at both scales. Pairwise turn-level agreement ranges from
86.48\% to 93.72\%, with Pearson correlation from 0.822 to 0.942 and Spearman
correlation from 0.714 to 0.943. We also collect blinded human-majority labels
for 60 randomly sampled outputs. Three-way human/model agreement is 85.0\%
(Fleiss' \(\kappa=0.705\)); human agreement with Qwen3.5-27B and
DeepSeek-V4-Flash is 90.0\% and 88.3\%, respectively (pairwise
\(\kappa=0.667\) for both).

\paragraph{LLM auditor prompt.}
The auditor receives the query or roleplay context, the response or visible
simulation transcript, and the frozen task-specific HR and CFR taxonomies. It
does not receive method labels, reward scores, reward traces, or training-stage
metadata when producing the reported HR/CFR labels. It returns compact JSON and
must prefer no label over a weak
or invented label:
\begin{tcblisting}{
    title=HR/CFR Auditor Prompt Excerpt,
    colback=white,
    colframe=black,
    width=\linewidth,
    listing only,
    listing options={basicstyle=\small\ttfamily,breaklines=true,columns=fullflexible},
    breakable,
}
Audit the response against the given task/context. Label only the predefined HR and CFR types. Do not invent new categories.

HR means the response contains a predefined hacking pattern that a rubric judge may over-reward without improving true task quality. CFR means the response has a predefined hard coverage failure that may be under-covered by the reward.

Use the task-specific HR and CFR type lists provided in the prompt. Mark a type only when it is meaningfully present and supported by visible evidence. Do not mark normal useful structure, necessary detail, appropriate roleplay narration, or ordinary low quality as HR/CFR. Prefer no over weak yes.

Return JSON only:
{
  "hack": "yes|no",
  "hack_types": ["..."],
  "coverage_failure": "yes|no",
  "coverage_failure_types": ["..."],
  "evidence": "<=25 words"
}
\end{tcblisting}

\paragraph{R-Var and Zero.}
For each query \(x\), let \(\mathcal{Y}_x\) be the grouped rollout responses and \(r(y)\) the scalar reward used by GRPO. R-Var is the average within-query reward variance:
\[
\mathrm{R\mbox{-}Var}=\frac{1}{|\mathcal{X}|}\sum_{x\in\mathcal{X}}
\mathrm{Var}\!\left(\{r(y):y\in\mathcal{Y}_x\}\right).
\]
Zero is the percentage of same-query groups whose reward range is indistinguishable under a small numerical tolerance:
\[
\mathrm{Zero}=\frac{1}{|\mathcal{X}|}\sum_{x\in\mathcal{X}}
\mathbf{1}\!\left[\max_{y\in\mathcal{Y}_x}r(y)-\min_{y\in\mathcal{Y}_x}r(y)\le \epsilon\right],
\]
with \(\epsilon=10^{-6}\) in our logged reward-health computation. Larger R-Var and lower Zero usually indicate more useful group-relative training signal, but high variance caused only by noisy penalties or score collapse is not treated as sufficient evidence of reward quality.

\subsection{Multi-Judge Benchmark Evaluation}
\label{app:benchmark_score_uncertainty}

We evaluate final-policy quality using GPT-5.6-Terra, DeepSeek-V4-Pro, and
GLM-5.2 as judges. Method identities, training rewards, reward traces, and
reward-system states are hidden from all judges. For each benchmark, the
reported score is the equal-weight average of the three judges on the common
set of instances successfully evaluated for every method--judge combination.

For WritingBench, each method produces responses to 1,000 test instructions.
Following the original benchmark protocol, a judge scores all five
query-specific criteria in a single request on a 1--10 scale. We average the
five criterion scores and multiply by ten to obtain the response-level score.
The common-success intersection contains 994 instructions; six instances
affected by provider-side filtering are excluded uniformly from all methods.

For CoSER, we evaluate 200 multi-turn roleplay simulations per method. Each
judge reads the complete dialogue and evaluates SC, AN, CF, and SQ in a single
request, preserving evidence that emerges across turns. For dimension \(d\), we
compute
\[
\begin{aligned}
u_d &= 100-5\max\left(0,\sum_j v_{d,j}-0.3T/4\right),\\
s_d &= \operatorname{clamp}(u_d,0,100),
\end{aligned}
\]
where \(v_{d,j}\) is the severity of flaw \(j\) and \(T\) is the number of
assistant turns. CoSER Overall is the mean of the four dimensions. The
common-success intersection contains 194 simulations; generation failures and
provider-filtered judge requests are excluded uniformly from all methods.

WritingBench judging uses temperature \(1.0\), top-\(p=0.95\), and a maximum of
4,096 output tokens. CoSER judging uses temperature \(0\) and a maximum of 2,048
output tokens. GPT-5.6-Terra uses its provider-default reasoning mode, while
reasoning is disabled for DeepSeek-V4-Pro and GLM-5.2. All evaluations were
conducted in September 2026.

\paragraph{Judge-specific results.}
Table~\ref{tab:judge_specific_quality} reports the results from each judge
before aggregation. \method{} obtains the highest WritingBench and CoSER Overall
score under every judge. The absolute scores and the magnitude of the roleplay
differences vary across evaluators, but the ranking of \method{} is consistent
across both benchmarks.

\begin{table*}[!tbp]
\centering
\small
\setlength{\tabcolsep}{4.2pt}
\begin{tabular}{lrrr|rrr}
\toprule
& \multicolumn{3}{c}{WritingBench} & \multicolumn{3}{c}{CoSER Overall} \\
\cmidrule(lr){2-4}\cmidrule(lr){5-7}
Method & GPT-5.6 & DeepSeek-V4 & GLM-5.2 & GPT-5.6 & DeepSeek-V4 & GLM-5.2 \\
\midrule
Qwen3-4B base & 49.815 & 60.638 & 54.229 & 57.075 & 66.960 & 58.692 \\
RLAIF & 48.992 & 63.384 & 56.451 & 56.944 & 68.189 & 54.585 \\
RaR & 49.374 & 63.489 & 56.779 & 56.885 & 68.568 & 54.548 \\
RLER & 49.746 & 62.771 & 55.994 & 56.245 & 67.141 & 55.477 \\
OpenRS & 49.857 & 63.127 & 56.296 & 53.967 & 67.544 & 54.281 \\
\textbf{\method{}} & \textbf{50.024} & \textbf{63.819} & \textbf{57.161}
& \textbf{62.242} & \textbf{71.387} & \textbf{63.400} \\
\bottomrule
\end{tabular}
\caption{\textbf{Judge-specific final-policy quality.} Each column reports
scores from one judge on the common-success subset used by the three-judge
aggregation.}
\label{tab:judge_specific_quality}
\end{table*}

\paragraph{Paired uncertainty estimates.}
We additionally perform paired bootstrap resampling over the common benchmark
instances. On WritingBench, \method{} has the highest mean score, with a
\(0.454\)-point margin over RaR. On CoSER, its improvements over both the base
policy and RaR remain positive throughout the paired bootstrap intervals. The
directional comparison shows the same pattern: \method{} outperforms RaR on 145
of the 194 matched roleplay simulations.

\begin{table}[!tbp]
\centering
\small
\setlength{\tabcolsep}{3.8pt}
\resizebox{\linewidth}{!}{%
\begin{tabular}{llrrr}
\toprule
Benchmark & Comparison & Mean \(\Delta\) & 95\% bootstrap interval & Win/Tie/Loss \\
\midrule
WritingBench & \method{} -- RaR & \(+0.454\) & \([-0.022,+0.918]\) & 502/29/463 \\
CoSER Overall & \method{} -- Base & \(+4.767\) & \([+3.537,+6.006]\) & 139/4/51 \\
CoSER Overall & \method{} -- RaR & \(+5.676\) & \([+4.487,+6.851]\) & 145/0/49 \\
\bottomrule
\end{tabular}%
}
\caption{Paired comparisons on the common-success benchmark subsets. Bootstrap
intervals use 20,000 paired resamples.}
\label{tab:benchmark_score_uncertainty}
\end{table}

\paragraph{Behavioral effects of fixed reward optimization.}
The roleplay results reveal that optimizing a fixed or partially adapted reward
system can improve selected qualities without improving the complete
interaction. Relative to the base policy, RaR increases SC by \(1.607\) points
and CF by \(1.991\) points, but decreases AN by \(2.420\) points and SQ by
\(4.812\) points. The resulting change is therefore not a uniform loss of
capability, but an uneven shift toward the dimensions most readily captured by
the training reward.

This imbalance is also visible directly in the generated dialogues. RLAIF,
RaR, RLER, and OpenRS produce longer conversations with substantially more
repeated phrases, repeated sentences, or length-truncated responses than the
base policy. Their judge rationales frequently identify circular dialogue,
scene stagnation, timeline contradictions, excessive literary elaboration, and
control of the user character. \method{} remains close to the base policy in
response length while substantially reducing repetition, suggesting that its
quality gains come from more coherent progression rather than increased
verbosity.

\begin{table}[!tbp]
\centering
\small
\setlength{\tabcolsep}{3.8pt}
\resizebox{\linewidth}{!}{%
\begin{tabular}{lrrrr}
\toprule
Method & Avg. chars & Repeated 5-gram & Repeated sentences & Length stop \\
\midrule
Qwen3-4B base & 15,883 & 22.9\% & 7.0\% & 8.4\% \\
RLAIF & 18,815 & 26.7\% & 10.9\% & 17.9\% \\
RaR & 24,173 & 31.1\% & 15.8\% & 27.0\% \\
RLER & 19,496 & 40.1\% & 23.1\% & 14.9\% \\
OpenRS & 20,649 & 30.2\% & 13.6\% & 37.4\% \\
\textbf{\method{}} & 15,634 & \textbf{12.2\%} & \textbf{1.1\%} & 9.6\% \\
\bottomrule
\end{tabular}%
}
\caption{\textbf{Behavioral characteristics of the evaluated roleplay
simulations.} Repetition and truncation statistics are computed directly from
the generated dialogues, independently of the quality judges.}
\label{tab:roleplay_behavioral_characteristics}
\end{table}

Together, the dimension-level and behavioral results show how a reward system
can become less reliable as the policy adapts to it: optimization continues to
strengthen reward-salient behavior while degrading qualities that the current
reward does not adequately represent. By evolving the reward system from
on-policy evidence, \method{} maintains more balanced supervision as the policy
distribution changes.

\section{Experimental Results and In-Depth Analysis}
\label{app:experimental_analysis}

\subsection{Detailed Generality and Ablation Results}
\label{app:ablation_details}

Table~\ref{tab:generality_detailed} expands the Qwen3-8B generality result with
judge-specific CoSER Overall scores and the three-judge average for each
dimension. Table~\ref{tab:downstream_ablation_detailed} provides the
dimension-level quality and reward-reliability audits for the downstream
ablations in Table~\ref{tab:downstream_component_ablation}.

\begin{table*}[!tbp]
\centering
\small
\setlength{\tabcolsep}{3.2pt}
\resizebox{\textwidth}{!}{%
\begin{tabular}{llrrrr|rrrr}
\toprule
& & \multicolumn{4}{c}{CoSER Overall} & \multicolumn{4}{c}{Three-judge dimensions} \\
\cmidrule(lr){3-6}\cmidrule(lr){7-10}
Method & GRM & GLM-5.2 & DeepSeek-V4 & GPT-5.6 & Mean & SC & AN & CF & SQ \\
\midrule
Qwen3-8B base & -- & 63.589 & 71.946 & 58.101 & 64.545 & 57.708 & 72.886 & 61.989 & 65.598 \\
RaR & Qwen3.5-27B & 61.861 & 70.748 & 53.524 & 62.044 & 55.975 & 69.700 & 60.783 & 61.719 \\
\textbf{\method{}} & Qwen3.5-27B & 63.687 & 74.698 & 60.726 & \textbf{66.370} & 65.732 & 69.978 & 65.943 & 63.829 \\
RaR & RewardAnything-8B & 61.720 & 72.581 & 53.982 & 62.761 & 56.942 & 70.430 & 61.364 & 62.309 \\
\textbf{\method{}} & RewardAnything-8B & 64.308 & 73.075 & 59.597 & \textbf{65.660} & 64.099 & 69.290 & 65.716 & 63.535 \\
\bottomrule
\end{tabular}%
}
\caption{\textbf{Detailed Qwen3-8B generality results.} Overall columns report
individual judges and their equal-weight mean; SC, AN, CF, and SQ report the
three-judge averages.}
\label{tab:generality_detailed}
\end{table*}

\begin{table*}[!tbp]
\centering
\small
\setlength{\tabcolsep}{3.0pt}
\resizebox{\textwidth}{!}{%
\begin{tabular}{lrrrrr|rr|rr}
\toprule
& \multicolumn{5}{c}{CoSER quality} & \multicolumn{2}{c}{Writing audits} & \multicolumn{2}{c}{Roleplay audits} \\
\cmidrule(lr){2-6}\cmidrule(lr){7-8}\cmidrule(lr){9-10}
Variant & Overall & SC & AN & CF & SQ & HR & CFR & HR & CFR \\
\midrule
Full \method{} & 65.676 & 63.615 & 69.880 & 63.381 & 65.829 & 6.5 & 19.9 & 5.25 & 10.0 \\
only criteria-level adaptation & 61.155 & 65.151 & 58.794 & 64.580 & 56.097 & 6.0 & 22.1 & 6.54 & 8.5 \\
Fixed final \rewarddag{} & 60.573 & 62.320 & 60.037 & 63.745 & 56.189 & 5.4 & 24.5 & 7.17 & 15.3 \\
\midrule
w/o candidate selection & 62.494 & 61.735 & 65.078 & 63.673 & 59.488 & 10.8 & 18.0 & 6.85 & 8.0 \\
w/o update rejection & 63.911 & 62.877 & 66.690 & 65.342 & 60.735 & 5.4 & 16.6 & 7.99 & 10.5 \\
w/o evolution memory / skills & 62.412 & 65.493 & 60.041 & 65.720 & 58.394 & 8.5 & 21.3 & 6.20 & 11.2 \\
\bottomrule
\end{tabular}%
}
\caption{\textbf{Detailed downstream ablations.} CoSER quality scores average
GPT-5.6-Terra, DeepSeek-V4-Pro, and GLM-5.2. HR and CFR retain the audit protocol
used in Table~\ref{tab:main_results}.}
\label{tab:downstream_ablation_detailed}
\end{table*}

\subsection{Candidate-State Trace Analysis}
\label{app:candidate_state_trace_analysis}

This subsection presents representative candidate states selected and not
selected during writing training, illustrating how comparative feedback shapes
the reward-system evolution path.
For selected states, we report the local two-step change in same-query R-Var
from the non-guard reward branch. This statistic describes the reward behavior
observed around each transition and is interpreted together with matched-case
comparative evidence, rather than as a standalone measure of downstream policy
improvement. Changes to flaw-penalty calibration can produce larger local
effects because they directly alter the separation of high-scoring failure
cases.

\begin{table}[!htbp]
\centering
\small
\setlength{\tabcolsep}{4.5pt}
\resizebox{\linewidth}{!}{%
\begin{tabular}{llrrr}
\toprule
Update family & Steps & Count & Total \(\Delta\) R-Var & Mean \(\Delta\) R-Var \\
\midrule
Structure / reward composition & 5, 15 & 2 & +0.000628 & +0.000314 \\
Query-rubric policy & 10, 25, 45, 65 & 4 & +0.001083 & +0.000271 \\
Guard calibration & 20, 35, 40, 55 & 4 & +0.001620 & +0.000405 \\
\bottomrule
\end{tabular}
}
\caption{Representative selected writing candidate states grouped by update family. The local
R-Var changes are two-step before/after changes from the non-guard
reward branch and are used as process diagnostics, not standalone downstream
quality gains. The positive/zero/negative counts are \(2/0/0\), \(4/0/0\), and
\(3/1/0\) for the three rows, respectively.}
\label{tab:selected_candidates_by_update_family}
\end{table}

\begin{table*}[!tbp]
\centering
\small
\setlength{\tabcolsep}{4pt}
\begin{tabular}{p{0.05\textwidth}p{0.19\textwidth}p{0.42\textwidth}p{0.25\textwidth}}
\toprule
Step & Update family & Candidate change & Selection evidence \\
\midrule
5 & Structure / query-rubric mechanism
& Added a \texttt{writing\_query\_judge} node so the reward could instantiate query-specific criteria from meta-rubrics instead of relying only on a global rubric.
& Matched replay verified the DAG path and established the query-level reward signal used by later updates. \\
\midrule
10 & Query-rubric policy
& Constrained query-rubric generation to produce criteria better matched to the current query and reduce flattening from generic criteria.
& Targeted replay showed mechanism activation and improved sampled same-query separation. \\
\midrule
15 & Structure / guard-style aggregation
& Added \texttt{writing\_artifact\_flaw\_penalty} and a multiplicative final aggregator so the artifact signal could only lower the base reward.
& Replay showed activation on high-score bad-tail patterns such as long templates, unsupported precision, and generic padding while preserving clean cases. \\
\midrule
20 & Guard calibration
& Calibrated the artifact guard to better trigger on word-count tags, mixed-script glitches, generic padding, and related observable artifacts.
& The previous guard under-activated; clean same-query replay showed improved separation. \\
\midrule
25 & Query-rubric reliability
& Strengthened query-rubric generation reliability so criteria are derived from the current query rather than reused as coarse generic dimensions.
& Comparative feedback supported the change, and execution checks passed. \\
\midrule
35 & Guard calibration
& Tightened generic-expansion and redundant-padding detection so section inflation and boilerplate are penalized when they lack query-specific substance.
& High-scoring long-template outputs still often received perfect guard scores; same-query replay showed improved sampled spread. \\
\midrule
40 & Guard calibration
& Tightened template-scaffold and meta-leakage detection for placeholders, word-count tags, scaffold shells, and similar artifacts.
& Selected as a narrow false-negative repair; local R-Var was approximately unchanged, but replay showed targeted activation without broad degradation. \\
\midrule
45 & Query-constraint policy
& Required query-rubric generation to include an explicit query-constraint fulfillment dimension covering format, length, structure, audience, and required content.
& Replay had no node errors and showed usable query-level spread. \\
\midrule
55 & Narrow guard rule
& Added a narrow rule that unresolved bracket placeholders must trigger the template-scaffold flaw rather than receiving a perfect guard score.
& Targeted replay fixed clear false negatives such as unresolved placeholder greetings without same-query degradation. \\
\midrule
65 & Query-rubric anti-template policy
& Tightened query-rubric generation to distinguish query-grounded completion quality from generic expansion, structural fluency, and template-like length.
& Replay had no node errors and successful query-rubric generation across sampled cases. \\
\bottomrule
\end{tabular}
\caption{Representative selected writing candidate states. Each row summarizes the update
family, the bounded reward-system change, and the comparative evidence that
supported its selection.}
\label{tab:selected_candidate_trace}
\end{table*}

\begin{table*}[!tbp]
\centering
\small
\setlength{\tabcolsep}{5pt}
\begin{tabular}{p{0.06\textwidth}p{0.39\textwidth}p{0.45\textwidth}}
\toprule
Step & Non-selected candidate change & Comparative evidence \\
\midrule
30 & Further tightened query-rubric generation to repair query-rubric degeneration and same-query flattening.
& Targeted replay showed lower same-query informativeness; broader replay was mixed and partly parse-confounded, so the candidate was not selected. \\
\midrule
50 & Further tightened the guard penalty for generic expansion.
& Replay reduced same-query standard deviation and range, indicating compression of sustainable reward signal; the candidate was not selected. \\
\midrule
60 & Made mixed-script artifacts trigger the guard more deterministically.
& Target activation was unstable and the penalty spilled over to non-template technical or plain-language cases. \\
\midrule
70 & Relaxed or rewrote query-rubric policy to reduce fallback and improve query-level informativeness.
& Candidate replay lowered the non-guard same-query signal and did not support selecting the new state. \\
\bottomrule
\end{tabular}
\caption{Representative non-selected writing candidates. These outcomes show
how comparative feedback redirects later search when observed reward behavior
indicates signal collapse, unstable triggering, parse-confounded behavior, or
penalty spillover.}
\label{tab:nonselected_candidate_trace}
\end{table*}

\subsection{Supplementary Reward-Variance Dynamics}
\label{app:reward_variance_dynamics}

\begin{figure}[!htbp]
\centering
\includegraphics[width=\linewidth]{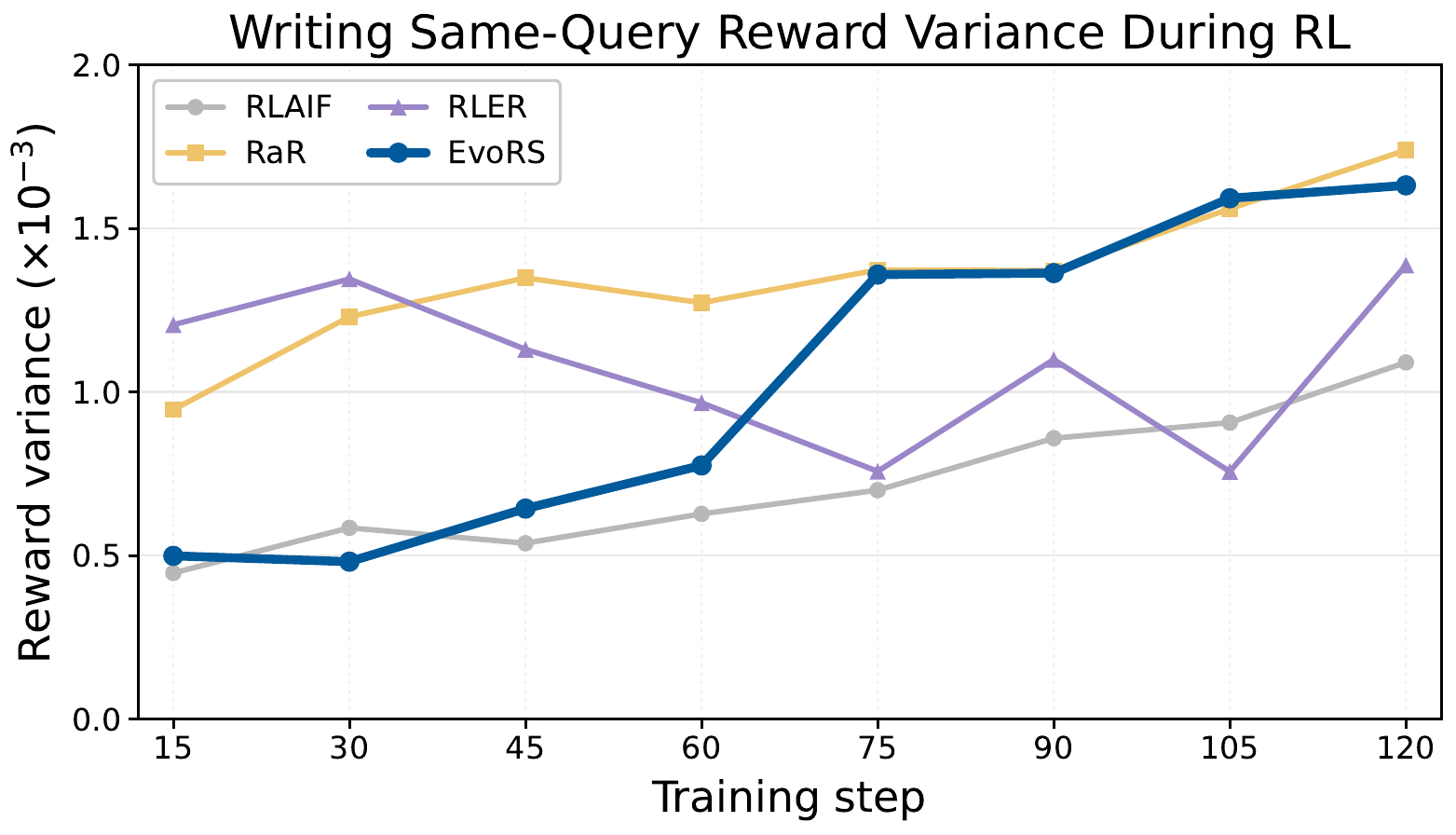}
\caption{Writing same-query reward variance during RL, computed over 15-step rollout windows from the non-guard continuous scoring branch. Higher variance can indicate greater within-query spread, but is interpreted together with zero-distinction and reward-ranking alignment in Section~\ref{sec:exp_signal_quality}.}
\label{fig:app_writing_reward_variance}
\end{figure}

\subsection{Reward-Signal Ranking Audit}
\label{app:reward_signal_ranking_audit}

Section~\ref{sec:exp_signal_quality} uses zero-distinction to test whether the
reward remains discriminative on on-policy rollout groups. This diagnostic
measures signal availability, but it does not by itself show that the reward
distinguishes responses in the correct direction. We therefore conduct a blind
list-wise ranking audit that compares reward-induced rankings with independent
preference rankings over the same response groups.

The independent ranking standard is method-agnostic and does not use
WritingBench criteria, method-specific rubrics, or reward-system rubrics.
Rankings are based on human-readable writing preferences that can be applied
uniformly across methods: whether the response satisfies the user request, fits
the requested format and audience, follows length and style constraints,
remains concrete and useful, avoids unsupported specificity or
self-certification, and is coherent as a finished writing deliverable. The same
ranking standard is used for every method and checkpoint.

For each sampled group, the packet contains one query and four same-query
candidate responses. Candidate identities are anonymized and shuffled.
Annotators do not observe method labels, reward scores, hidden mappings,
rollout source paths, or the reward ranking. Hidden mappings and logged reward
scores are used only after all rankings for a packet are completed.

Each packet is independently ranked by a three-model LLM judge panel:
GPT-5.4, DeepSeek-V4-Pro, and Kimi-K2.5. We aggregate the three rankings with
Borda voting to obtain a consensus ranking, and we report inter-judge pairwise
agreement as a stability diagnostic. This design reduces sensitivity to any
single judge trajectory and lets us separate
reward-alignment results from annotation instability. Human review is used for
packet validation and calibration: human annotators provide an initial
preference prior, inspect packet construction and blinding, and check a
20-group manually ranked subset against the LLM consensus ranking.

\begin{table}[!htbp]
\centering
\small
\setlength{\tabcolsep}{5pt}
\begin{tabular}{p{0.48\linewidth}cp{0.23\linewidth}}
\toprule
Reliability check & Groups & Pairwise agreement \\
\midrule
Three-model LLM judge panel & 480 & 0.8408 \\
Human vs. LLM consensus & 20 & 0.71 \\
\bottomrule
\end{tabular}
\caption{Reliability checks for the blind list-wise ranking audit. The
three-model value averages inter-judge pairwise agreement over all
method/checkpoint groups. The human-vs-LLM row reports the manually checked
20-group subset.}
\label{tab:ranking_audit_reliability}
\end{table}

\begin{figure*}[!tbp]
\centering
\begin{minipage}{0.48\textwidth}
\centering
\includegraphics[width=\linewidth]{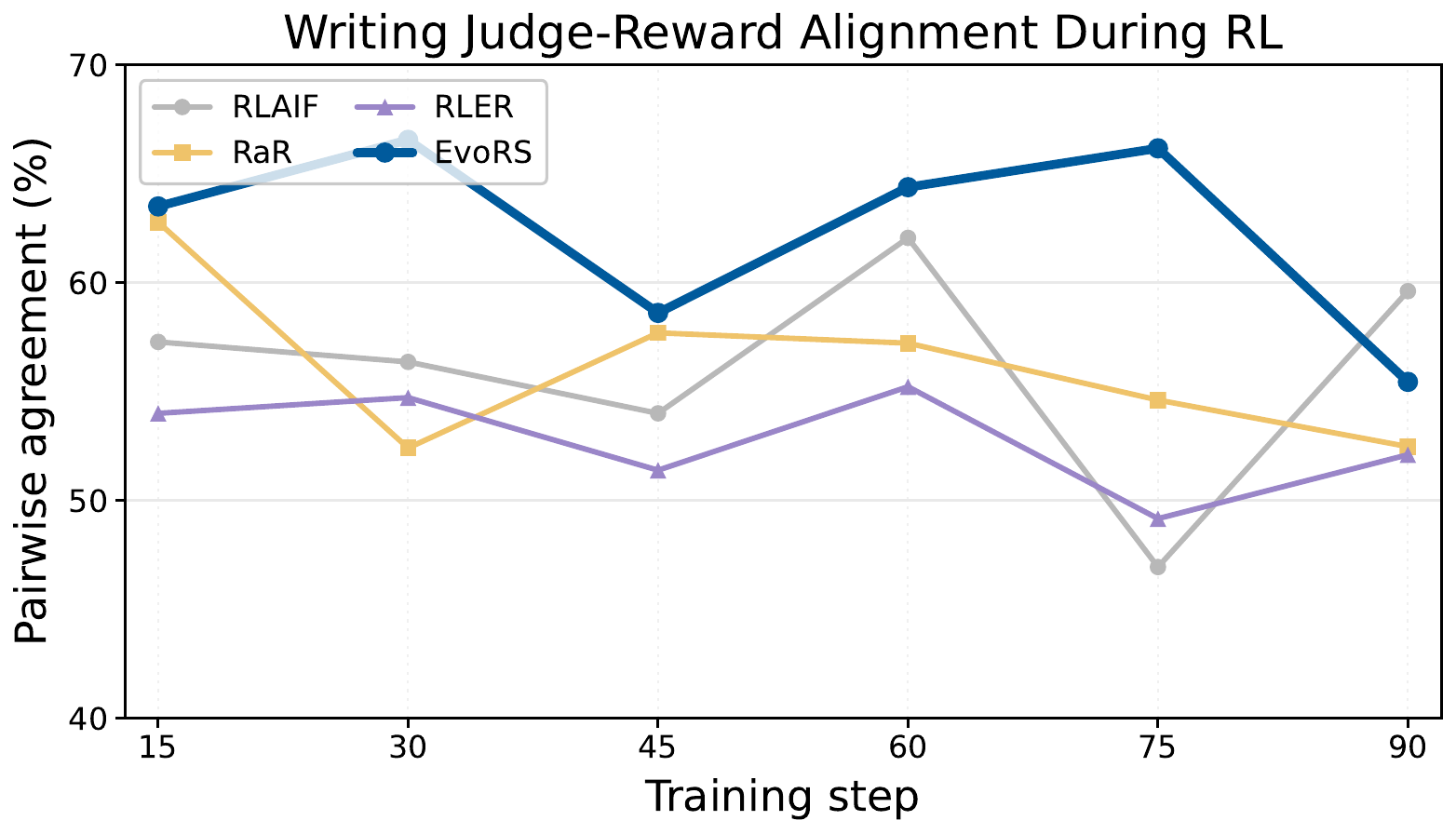}
\end{minipage}
\hfill
\begin{minipage}{0.48\textwidth}
\centering
\includegraphics[width=\linewidth]{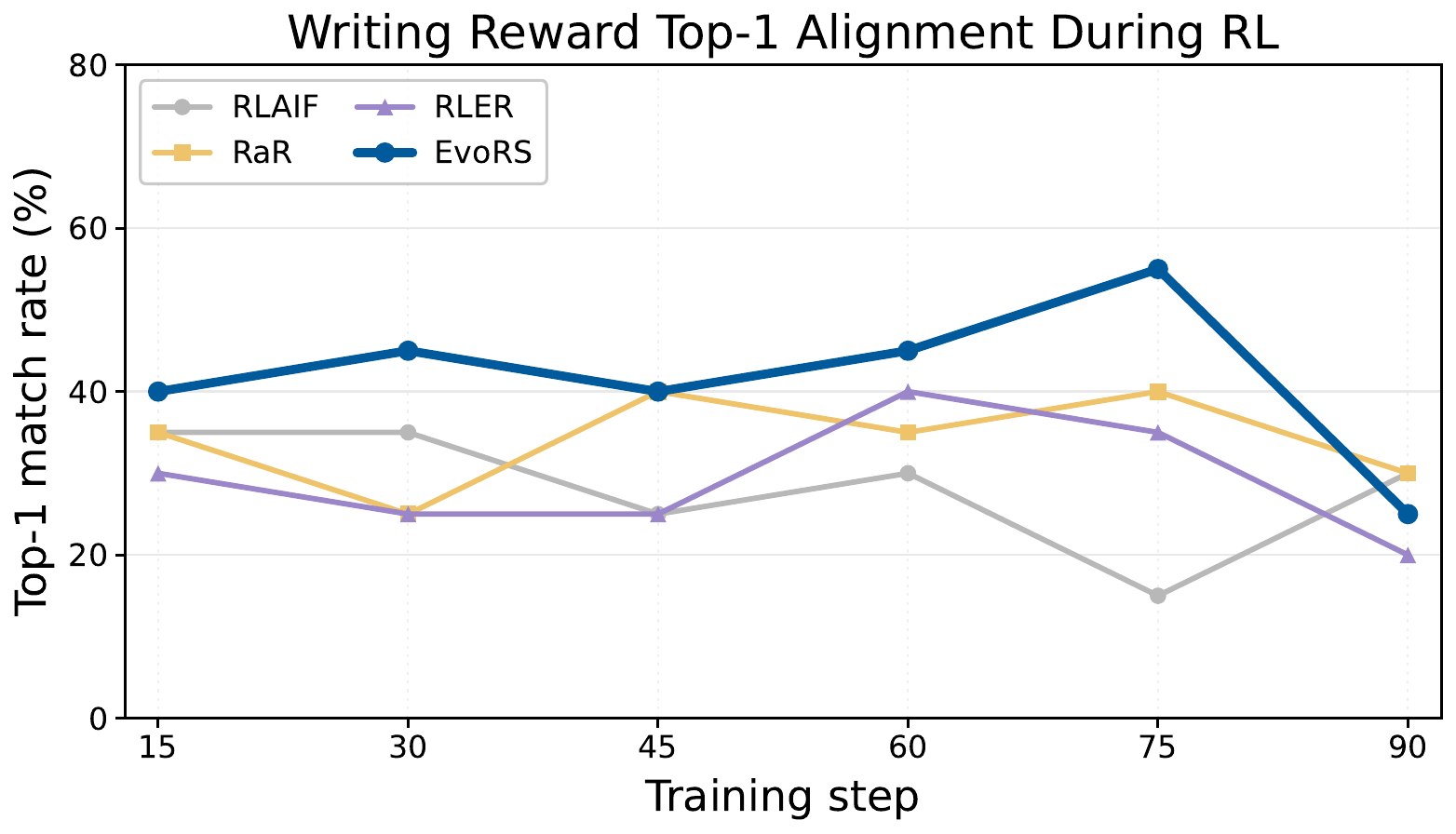}
\end{minipage}
\vspace{0.6em}

\begin{minipage}{0.48\textwidth}
\centering
\includegraphics[width=\linewidth]{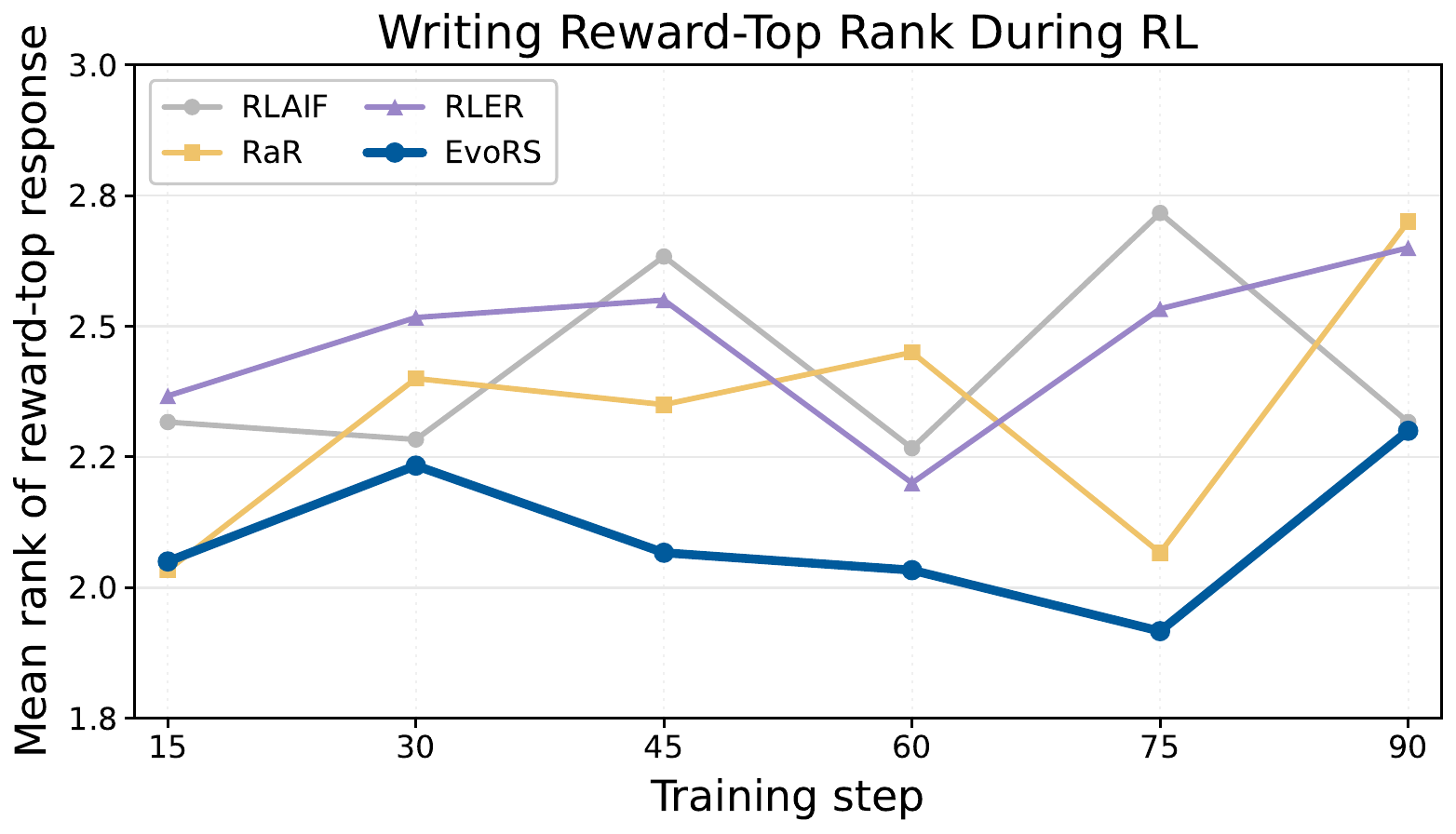}
\end{minipage}
\hfill
\begin{minipage}{0.48\textwidth}
\centering
\includegraphics[width=\linewidth]{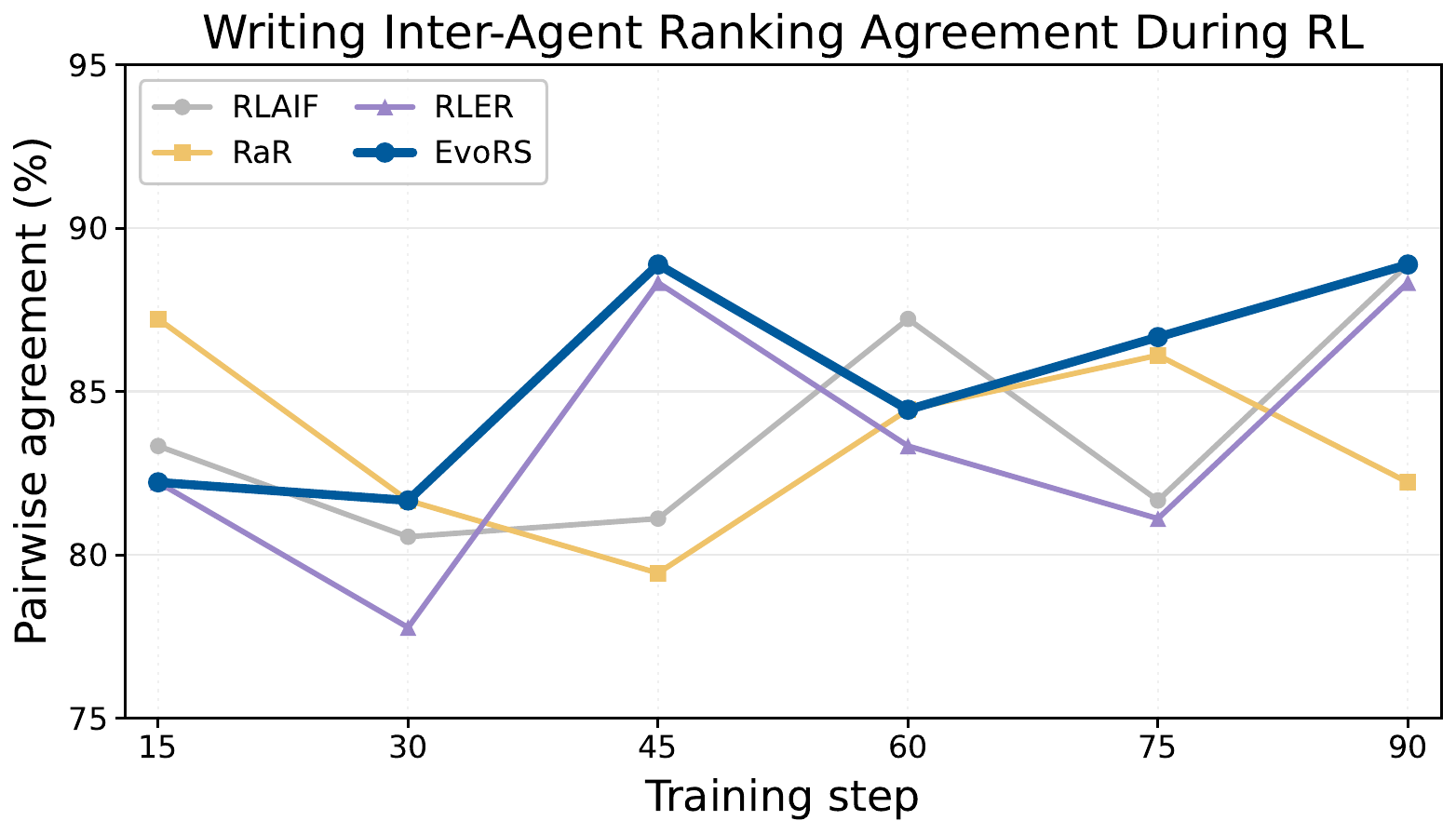}
\end{minipage}
\caption{Supplementary diagnostics for the blind reward-ranking audit. These
curves complement the main consensus pairwise ranking agreement curve in
Figure~\ref{fig:training_time_dynamics}(d).}
\label{fig:app_reward_ranking_audit_aux}
\end{figure*}

The main metric is consensus pairwise ranking agreement. A four-response
ranking induces six pairwise preferences. We compare the six preferences from
the independent consensus ranking with the six preferences induced by reward
scores, ignoring reward ties, and average the agreement over groups. We
additionally report auxiliary diagnostics in
Figure~\ref{fig:app_reward_ranking_audit_aux}. Judge-vs-reward pairwise
agreement averages each LLM judge's ranking against the reward ranking
before consensus. Top-1 match measures whether the reward-selected best
response is also ranked first by the independent judge. Reward-top mean rank
measures where the reward-selected best response appears in the independent
ranking; lower is better. Inter-judge agreement measures the intrinsic
reliability of the list-wise annotation task.

\subsection{Case Studies}
\label{app:case_studies}

\paragraph{Writing guard-style penalty repair.}
This case supports the main-text hack-rate analysis. The frozen evaluation taxonomy in Appendix~\ref{app:hack_aware_audit} was not exposed to the designer. The agent instead inspected on-policy rollout evidence and reward traces, then diagnosed overlapping failure modes in its own terms. In the writing run used in Figure~\ref{fig:training_time_dynamics}(a), a guard-style penalty path was introduced at step 15 and calibrated at the annotated step-20 transition. Before the structural update, the active graph contained a continuous Global Rubric Node, a Query-Level Rubric Node, and a terminal weighted-sum Composition Operator:
\begin{quote}\small
\texttt{writing\_global\_judge} \(\rightarrow\) \texttt{writing\_query\_judge} \(\rightarrow\) \texttt{final\_reward}, where \texttt{final\_reward} was a weighted sum over the global and query-level scores.
\end{quote}
The designer diagnosed concrete reward-hacking evidence: many rollout responses contained word-count tags, placeholders, or template residue, and several artifact-heavy responses still received high rewards. These observations overlap with the evaluation-time families of fabricated authority, self-certification, and one-dimension over-optimization, but were derived from rollout diagnostics rather than from the external auditor labels. Among the candidate repair directions, the selected state added a bounded integrity side node after the query judge and marked it as \texttt{role=guard} in the final aggregator. The side node covers artifact-free output, specificity/authenticity, and anti-template behavior. The final reward can be viewed as
\[
r_{\mathrm{final}}=h_{\mathrm{base}}\cdot h_{\mathrm{guard}},
\]
where \(h_{\mathrm{base}}\) is the weighted positive score from the global and query-level nodes, and \(h_{\mathrm{guard}}\le 1\) can only reduce rewards for detected artifacts.

Later updates refined the same guard-style path rather than introducing the frozen evaluation labels. A step-90 probe attempted to switch the side-node scoring mode to \texttt{flaw\_penalty}, but replay revealed an operator-specific failure: empty flaw extraction mapped to an always-pass score, so this probe was treated as diagnostic evidence rather than a successful repair. Subsequent selected updates at steps 95--115 made bounded local updates to the guard-role node and grounding items, targeting mixed-script glitches, placeholders, explicit word-count tags, and dense unsupported numeric/entity specificity. Replay evidence showed activation on targeted bad cases while preserving strong clean examples in sampled checks. This case therefore shows a structure-level guard-style penalty repair and later calibration, not merely a criteria-only dynamic rubric rewrite.

\paragraph{Roleplay coverage repair.}
\label{app:on_policy_coverage_case}
This case supports the main-text on-policy coverage analysis. In a representative roleplay evolution step, the designer diagnosed that the query-side roleplay reward under-covered latest-turn consequence tied to visible scene-state changes. Several high-scoring responses were long and polished, but relied on decorative \texttt{<role-think>} padding, transplantable taunts, or unsupported specifics instead of concrete scene movement. A sampled same-query Morgan group was nearly flat, with score range about \(0.01\), indicating that the reward did not separate materially different continuations.

The selected candidate updated only \texttt{roleplay\_query\_rubric}. It strengthened criteria for latest-turn consequence, substantive content inside required roleplay blocks rather than tag compliance alone, and grounded progression with economy. It also demoted long private planning, decorative inner monologue, catchphrase or persona-name stuffing, and unsupported invented specifics. This is a positive-side coverage/calibration repair rather than a hack-only guard. Comparative feedback supported this candidate: mean score changed from \(0.6313\) to \(0.6390\), within-group standard deviation from \(0.0765\) to \(0.0844\), and within-group range from \(0.1969\) to \(0.2177\). The case supports the main-text claim that the designer can repair missing on-policy quality dimensions, not only penalize obvious hacking artifacts.

\end{document}